\documentclass{article}

\usepackage[dblblindworkshop, final]{neurips_2025}
\usepackage[utf8]{inputenc} % allow utf-8 input
\usepackage[T1]{fontenc}    % use 8-bit T1 fonts
\usepackage{hyperref}       % hyperlinks
\usepackage{url}            % simple URL typesetting
\usepackage{booktabs}       % professional-quality tables
\usepackage{amsfonts}       % blackboard math symbols
\usepackage{nicefrac}       % compact symbols for 1/2, etc.
\usepackage{microtype}      % microtypography
\usepackage{xcolor}         % colors

\usepackage{amsmath}
\usepackage{amssymb}
\usepackage{graphicx}
\usepackage{array}
\usepackage{caption}
\usepackage{tabularx}
\usepackage{multirow}
\usepackage{listings}
\usepackage{listingsutf8}
\usepackage{arydshln}
\usepackage{subcaption}
\usepackage{pifont}
\usepackage{tcolorbox}
\usepackage{pgfplots}
\pgfplotsset{compat=1.18}
\usepackage{siunitx}
\usepgfplotslibrary{colorbrewer}

\newcolumntype{Y}{>{\raggedright\arraybackslash}X}
\newcommand{\pmark}{{\raisebox{0.15ex}{\large$\blacktriangle$}}} 
\DeclareRobustCommand{\uparrowgreen}{%
  \textcolor{green!60!black}{\raisebox{0.15ex}{\scalebox{0.9}{$\uparrow$}}}%
}
\DeclareRobustCommand{\downarrowred}{%
  \textcolor{red!70!black}{\raisebox{0.15ex}{\scalebox{0.9}{$\downarrow$}}}%
}
\definecolor{nicegreen}{RGB}{34,139,34} %custom color i am using in the rest of the article. keep it

\title{Modality-Swap Distillation: Rendering Textual Reasoning into Visual Supervision}
\workshoptitle{Foundations of Reasoning in Language Models}

\author{%
  Boammani Aser Lompo \thanks{Equal contribution, random ordering.} \\
École de Technologie Supérieure\\
Montreal, Canada \\
\texttt{boammani.lompo.1@ens.etsmtl.ca} \\
\And
Marc Haraoui \footnotemark[1]\\
\texttt{mharaoui.pro@gmail.com}\\}

\begin{document}

\maketitle

\begin{abstract}
Visual reasoning over structured data such as tables is a critical capability for modern vision-language models (VLMs), yet current benchmarks remain limited in scale, diversity, or reasoning depth, especially when it comes to rendered table images. Addressing this gap, we introduce \textbf{Visual-TableQA}, a large-scale, open-domain multimodal dataset specifically designed to evaluate and enhance visual reasoning over complex tabular data. Our generation pipeline is \textbf{modular, scalable, and fully autonomous}, involving multiple reasoning LLMs collaborating across distinct roles: generation, validation, and inspiration. \textbf{Visual-TableQA} comprises 2.5k richly structured LaTeX-rendered tables and 9k reasoning-intensive QA pairs, all produced at a cost of under \$100. To promote diversity and creativity, our pipeline performs \textbf{multi-model collaborative data generation} via \textbf{cross-model prompting (‘inspiration’)} and LLM-jury filtering. Stronger models seed layouts and topics that weaker models elaborate, collectively distilling diverse reasoning patterns and visual structures into the dataset. Empirical results show that models fine-tuned on \textbf{Visual-TableQA} generalize robustly to external benchmarks, outperforming several proprietary models despite the dataset’s synthetic nature. The full pipeline and resources are publicly available in our \href{https://github.com/AI-4-Everyone/Visual-TableQA}{GitHub repository}.
\end{abstract}

\section{Introduction}
\begin{figure}[htb]
    \centering
    \includegraphics[width=\textwidth]{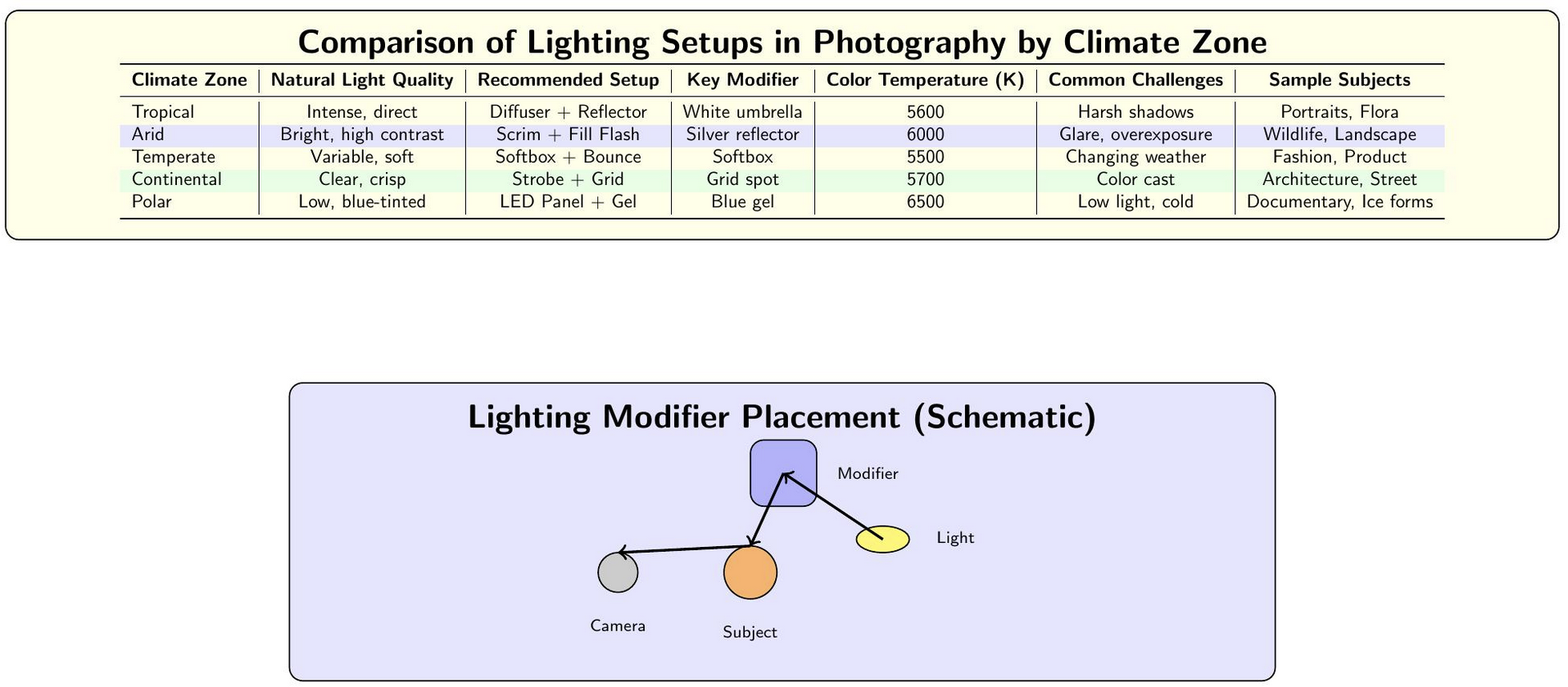}
    \caption{Sample question in our benchmark}
    \label{fig:qa_sample}
\end{figure}

Vision-language models (VLMs) have significantly advanced in recent years, achieving remarkable performance in various tasks involving visual and textual inputs. Despite these advancements, complex reasoning tasks, especially those requiring deep comprehension of tabular data structures, continue to pose significant challenges. Table complexity can manifest in various ways, including structural layout, information density, and the diversity of visual components such as the integration of diagrams. The more complex a table is, the more it lends itself to challenging reasoning tasks, requiring advanced cognitive abilities to extract relevant information and perform multi-step logical analysis. For example, the table in Figure~\ref{fig:qa_sample} exemplifies this complexity through its use of multirow cells, integrated diagrams, and color encoding. Answering the question requires the VLM to interpret information across all cells and perform a sequence of reasoning steps.

Existing table-based QA datasets predominantly fall into two categories: \textit{(i)} those represented purely in textual format—such as WikiTableQuestions~\citet{pasupat-liang-2015-compositional}, HybridQA~\citet{chen-etal-2020-hybridqa}, and AIT-QA~\citet{katsis-etal-2022-ait}—which bypass the challenges of visual layout interpretation; and \textit{(ii)} those that lack diversity in visual layouts, visual complexity, and reasoning depth due to being domain-specific (e.g., TAT-DQA~\citet{zhu2022towards}), or having standardized queries (e.g., TableVQA-Bench~\citet{kim2024tablevqa}), or highly technical in nature (e.g., Table-VQA~\citet{AgDeTQA}). This second datasets category typically rely on a limited set of layout templates and involve relatively simple visual tasks or basic QA scenarios, falling short of the complexity required for thorough evaluation and advancement of reasoning capabilities. More recent efforts—such as ChartQA~\citet{masry2022chartqa}, ReachQA~\citet{hedistill}, and MATH-Vision~\citet{wang2024measuring}—have aimed to address the need for open-domain coverage, incorporating more diverse visual features, varied question types, and deeper reasoning challenges. However, these datasets primarily focus on charts and function plots, overlooking tables—and with them, an entire dimension of informational structure and layout diversity. An extensive comparison of diverse chart and table datasets is provided in Table~\ref{tab:dataset-comparison}.

Inspired by ReachQA’s Code-as-Intermediary Translation (CIT)—a technique that translates chart images into textual representations while faithfully preserving visual features—we introduce \textbf{Visual-TableQA}, a novel synthetic, multimodal, and open-domain dataset tailored to enhance reasoning capabilities through complex table-based question-answering tasks. {Visual-TableQA} capitalizes on the ability of reasoning-oriented LLMs to generate intricate LaTeX tables, thus significantly reducing costs and eliminating the need for extensive manual annotations. This modality-swap makes it possible for LLMs to invest their textual reasoning ability into visual image in order to improve visual understanding and reasoning. {Visual-TableQA} emphasizes structural reasoning over domain knowledge. Each entry couples a rendered table image with a complex, visually grounded reasoning task. Tasks require interpreting visual layout cues such as cell alignment, hierarchical headers, merged cells, or embedded symbolic content—emulating real-world documents where visual context is essential for correct interpretation. The dataset contains 2.5k reasoning-intensive tables and 9k QA pairs crafted to assess both information extraction and multi-step reasoning capabilities, all generated at a cost of under \$100. The entire dataset has been validated using a committee of high-performing reasoning LLMs, the ROSCOE step by step reasoning score ~\citet{golovnevaroscoe}, and a sample of 800 QA pairs has undergone manual verification by human annotators. In contrast to previous synthetic datasets, {Visual-TableQA} is less guided in its generation process, allowing for more diversity and creativity in both table complexity (e.g., structural layout, information density, visual component variety) and the design of QA pairs explicitly crafted to challenge visual reasoning skills. We evaluated a broad range of VLMs, from lightweight models to state-of-the-art architectures, and benchmarked their performance against existing datasets. The results show that most VLMs continue to struggle with table understanding.

In sum, our main contributions are:\textit{(i)} a high-quality, visually diverse, and open-domain dataset for table-based reasoning; \textit{(ii)} an LLM-driven, low-cost generation pipeline using cross-model inspiration; \textit{(iii)} an empirical analysis comparing Visual-TableQA to existing table and chart datasets; \textit{(iv)} an extensive evaluation of open and proprietary VLMs, showing performance gains after finetuning. Our dataset and code are publicly available at \url{https://github.com/AI-4-Everyone/Visual-TableQA}.

\begin{table}[htb]
\caption{Comparison of existing chart and table datasets across data, Q\&A, and dataset properties. Abbreviations: Repr=Representation, Vis= Visual, Comp= Complexity, Temp = Template, Refer = Reference, Rat = Rational, Synth= Synthetic, Scal = Scalable. Cells marked with \pmark indicate mixed attributes (e.g., partially template-based; scalable Q\&A but non-scalable chart data)}
\centering
\scriptsize
\begin{tabularx}{\textwidth}{Y >{\centering\arraybackslash}p{1cm} *{2}{>{\centering\arraybackslash}p{0.5cm}} *{2}{>{\centering\arraybackslash}p{0.5cm}} *{2}{>{\centering\arraybackslash}p{0.5cm}} >{\centering\arraybackslash}p{0.5cm} >{\centering\arraybackslash}p{1cm} >{\centering\arraybackslash}p{0.5cm}}
\toprule
\multirow{5}{=}{\centering\arraybackslash\textbf{Datasets}}
& \multicolumn{4}{c}{\textbf{Data Properties}}
& \multicolumn{3}{c}{\textbf{Q\&A Properties}}
& \multicolumn{3}{c}{\textbf{Dataset Properties}}\\
\cmidrule(lr){2-5}\cmidrule(lr){6-8}\cmidrule(lr){9-11}
& \# Layouts/ \# Topics& Type & Data Repr. & Vis. Comp. & Temp. Free & Vis. Refer. & Rat. Annot. & Synth. & \#Samples / \#QA & Scal.\\
\midrule
WikiTableQuestions~\citep{pasupat-liang-2015-compositional} & -- & Table & Text & \ding{55} & \ding{55} & \ding{55} & \ding{55} & \ding{55} & 2.1k/22k & \ding{55}\\
HybridQA~\citep{chen-etal-2020-hybridqa} & -- & Table & Text & \ding{55} & \ding{51} & \ding{55} & \ding{55} & \ding{55} & 13k/70k & \ding{55}\\
AIT-QA~\citep{katsis-etal-2022-ait} & --/1 & Table & Text & \ding{55} & \ding{51} & \ding{55} & \ding{55} & \ding{55} & 116/515 & \ding{55}\\
TAT-DQA~\citep{zhu2022towards} & --/1 & \pmark & Image & \ding{55} & \ding{51} & \ding{51} & \ding{51} & \ding{55} & 2.5k/16.5k & \ding{55}\\
Table-VQA~\citep{AgDeTQA} & --/-- & Table & Image  & \ding{55} & \ding{51} & \ding{51} & \ding{51} & \ding{51} & 16.4k/82.3k & -- \\
TableVQA-Bench~\citep{kim2024tablevqa} & 11/4 & Table & Image  & \ding{55} & \ding{51} & \ding{51} & \ding{55} & \pmark & 894/1.5k & \pmark \\
ChartQA~\citep{masry2022chartqa} & 3/15 & Chart & Image & \ding{55} & \ding{51} & \ding{51} & \ding{55} & \ding{55} & 21.9k/32.7k & \ding{55} \\
DocVQA~\citep{mathew2020docvqa} & 20/5 & \pmark & Image & \ding{55} & \ding{51} & \ding{51} & \ding{55} & \ding{55} & 12.7k/50k & \ding{55}\\
MultiModalQA~\citep{talmor2021multimodalqa} & 16/$\infty$  & \pmark & Image & \ding{55} & \ding{55} & \ding{51} & \ding{55} & \pmark & 29,918 & \ding{55} \\
MATH-Vision~\citep{wang2024measuring} & --/16 & \pmark & Image & \ding{51} & \ding{51} & \ding{51} & \ding{55} & \ding{55} & 3k/3k & \ding{55} \\
REACHQA~\citep{hedistill} & 32/$\infty$ & Chart & Image & \ding{51} & \ding{51} & \ding{51} & \ding{51} & \ding{51} & 3.7k/22k & \ding{51} \\
\addlinespace[5pt]
\textbf{Visual-TableQA (ours)} &  /$\infty$ & Table & Image & \ding{51} & \ding{51} & \ding{51} & \ding{51} & \ding{51} & 2.5k/ 9k & \ding{51}\\
\bottomrule
\end{tabularx}
\label{tab:dataset-comparison}
\end{table}

\section{Visual-TableQA Dataset}

\begin{table}[htb]
\centering
\scriptsize
\caption{Model performance on the test sets of four benchmarks: {ChartQA}, {ReachQA}, {MATH-Vision}, {Visual-TableQA}, and {Visual-TableQA-CIT}. Visual-TableQA-CIT is the variant of our dataset where tables are represented in LaTeX code form rather than as rendered images. The ReachQA score is reported as the average across its two evaluation splits: \textit{Reasoning} and \textit{Recognition}. The values in {\color{blue}blue} are from our own evaluation using the LLM jury, while the remaining values are taken from model authors or official leaderboards/model cards. When a fine-tuned model achieves better performance, the result is annotated with \uparrowgreen; if the performance worsens, it is marked with \downarrowred. The best performance for each model variants and task is in \textbf{bold}.}

\renewcommand{\arraystretch}{1.3}
\begin{tabularx}{\textwidth}{p{4.4cm} @{\hspace{1em}} *{2}{>{\arraybackslash}p{1.3cm}} *{3}{>{\arraybackslash}X}}
\toprule
\textbf{Models} & \textbf{ChartQA} & \textbf{ReachQA} & \textbf{MATH-Vision}$_{FULL}$ & \textbf{Visual-TableQA} & \textbf{Visual-TableQA-CIT}\\
\midrule
\multicolumn{6}{c}{\textbf{Baseline}} \\
\hline
 Human & -- & 74.85 & 68.82 & -- & --\\
\hline
\multicolumn{6}{c}{\textbf{Proprietary VLMs}} \\
\hline
GPT-4o & 85.7 & 53.25 & 30.39 &\textcolor{blue}{78.5} &\textcolor{blue}{91.0}\\
GPT-4o mini & 77.52 & 40.35 & 28.85 &\textcolor{blue}{67.0} &\textcolor{blue}{80.27}\\
Gemini 2.5 Flash & \textcolor{blue}{84.64} &\textcolor{blue}{56.97} & 41.3 &\textcolor{blue}{85.72}&\textcolor{blue}{92.3}\\
Gemini 2.5 Pro &\textcolor{blue}{85.73} &\textcolor{blue}{61.87} & \textbf{73.3} &\textcolor{blue}{78.62(86.63)} &\textcolor{blue}{86.27}\\
Claude 3.5 Sonnet & $\textbf{90.8}^*$ & \textbf{63} & 32.76 & \textbf{\textcolor{blue}{82.46}} & \textbf{\textcolor{blue}{88.5}}\\
\hline
\multicolumn{6}{c}{\textbf{Open-Source VLMs}} \\
\hline
Llama 4 Maverick 17B-128E Instruct & $85.3^*$ & \textcolor{blue}{47.98 (49.75)} &\textbf{\textcolor{blue}{45.89}} & \textcolor{blue}{80.75} &\textcolor{blue}{87.0}\\
Mistral Small 3.1 24B Instruct & $86.24^*$ & \textcolor{blue}{42.45} & \textcolor{blue}{32.45} &\textcolor{blue}{73.2} &\textcolor{blue}{80.25}\\
Qwen2.5-VL-32B-Instruct &\textbf{\textcolor{blue}{79.75}} & \textbf{\textcolor{blue}{49.5}} & 38.1 & \textbf{\textcolor{blue}{80.45}} & \textbf{\textcolor{blue}{81.93}}\\
Qwen2.5-VL-7B-Instruct & $87.3^*$ & \textcolor{blue}{49.23} & 25.1 & \textcolor{blue}{71.35} &\textcolor{blue}{--}\\
\hline
\multicolumn{6}{c}{\textbf{Finetuned VLMs}} \\
\hline
Qwen2.5-VL-7B-Instruct + Visual-TableQA & \textbf{\textcolor{blue}{84.52}\uparrowgreen} & \textcolor{blue}{\textbf{60.95} \uparrowgreen} & \textbf{\textcolor{blue}{49.77 \uparrowgreen}} & \textbf{\textcolor{blue}{82.98 \uparrowgreen}} &\textcolor{blue}{N/A}\\
Qwen2.5-VL-7B-Instruct + ReachQA & \textcolor{blue}{77.59} \downarrowred & \textcolor{blue}{55.75} \uparrowgreen & \textcolor{blue}{48.57} \uparrowgreen & \textcolor{blue}{60.68 (56.13) \downarrowred} &\textcolor{blue}{N/A}\\
\bottomrule

\multicolumn{6}{p{13.6cm}}{\footnotesize * Performance metrics are measured using \textit{Relaxed Accuracy}, which allows for small numerical deviations in the predicted answers. We assume that this accuracy inflates the actual accuracy by at least 5\%. This margin is subtracted when selecting the best-performing results, which are shown in \textbf{bold}.}

\end{tabularx}
\label{tab:bench-results}
\end{table}

Unlike previous datasets that rely heavily on textual input or handcrafted annotations, Visual-TableQA leverages a scalable generation pipeline rooted in LaTeX-rendered table images, automated reasoning task creation, and LLM-based evaluation. This strategy enables high diversity and reasoning depth while keeping annotation costs minimal, totaling under \$100  using a combination of open-access APIs and limited usage tiers. In this section, we describe our LaTeX-based table encoding~\ref{sec:encoding}, the data generation pipeline~\ref{sec:pipeline}, and the quality assurance process~\ref{sec:quality}.

\subsection{\label{sec:encoding}Modality-Swap: Table Representation in LaTeX}
Our approach is inspired by ~\citet{hedistill}, which demonstrated that state-of-the-art VLMs can reason about visual content even in the absence of explicit visual input. Building on this insight, and leveraging the strong coding capabilities of reasoning-oriented language models across multiple programming languages, we chose to use an intermediate representation of tables in LaTeX rather than directly generating rendered table images. This strategy enables the generation of complex visual tables as compact LaTeX code—typically around 100 lines per table—drastically reducing the cost of generation by minimizing the number of output tokens required in API calls. We refer to this as modality-swap: LLMs invest their textual reasoning ability (programming and analysing programs) into visual images in order to improve visual understanding and reasoning. An ablation study of the modality-swap assumption is reported in Table~\ref{tab:bench-results}. We observe that most state-of-the-art models perform better when the same query is provided in textual form rather than as a visual input. This suggests both room for improvement in visual reasoning and the potential for transferring textual reasoning capabilities to the visual modality.

Our observations (Table~\ref{tab:generation_accuracy}) align with those of ~\citet{kale2025texpert}, who reported that LLMs struggle with LaTeX generation—particularly as task complexity increases, leading to a notable drop in accuracy. Table~\ref{tab:generation_accuracy} presents the performance of various models in generating LaTeX tables that compile without errors.

\begin{table}[htb]
\centering
\scriptsize
\caption{Percentage of successful LaTeX compilations for various models. Each accuracy is computed from at least 500 generated samples. The Adjust column indicates the level of manual correction needed to make the table look good: Low means minimal or no adjustments, Medium corresponds to 3–5 required fixes, and High indicates more than 5 adjustments were necessary. The tables generated by DeepSeek-R1-Distill-Qwen-32B never compiled.}
\begin{tabularx}{\textwidth}{X *{2}{>{\centering\arraybackslash}p{0.5cm}} @{\hspace{4em}} X *{2}{>{\centering\arraybackslash}p{0.5cm}}}
\toprule
\textbf{Model} & \textbf{Acc. (\%)} & \textbf{Adjust} & \textbf{Model} & \textbf{Acc. (\%)} & \textbf{Adjust} \\
\midrule
Llama 4 Maverick 17B-128E Instruct~\citet{meta2025_llama4_maverick_17b_128e_instruct}& 69 & High & DeepSeek-R1-Distill-Qwen-32B~\citet{deepseek2025_r1_distill_qwen_32b} & 0.0 & --\\
Gemini 2.0 Flash~\citet{google2025_gemini_2_0_flash} & 65.7 & Low & DeepSeek-R1T-Chimera~\citet{tng_technology_consulting_gmbh_2025} & 43.4 & Medium\\

Gemini 2.5 Flash~\citet{google2025_gemini_2_5_flash_model_card} & 43 & Medium & Claude Sonnet 4~\citet{anthropic2025_claude_sonnet_4_system_card} & 56 & Low\\

Gemini 2.5 Pro~\citet{google2025_gemini_2_5_pro_model_card} & 19.6 & Low & Claude 3.5 Haiku~\citet{anthropic2024_claude_3_5_haiku_addendum} & 64.4 & Low\\

GPT-4.1~\citet{openai2025_gpt4_1} & 41.5 & Low & Grok 3 Beta~\citet{xai2025_grok_3_beta} & 47.3 & Low\\

Qwen3-30B-A3B~\citet{qwen2025_qwen3_30b_a3b} & 69.4 & Low & Reka Flash 3~\citet{reka2025_flash3_model_card} & 19.3 & Medium\\

Qwen-QwQ-32B~\citet{qwen2025_qwq_32b} & 38.2 & Low\\

\bottomrule
\end{tabularx}
\label{tab:generation_accuracy}
\end{table}

\subsection{\label{sec:pipeline}Data Generation Pipeline}
\begin{figure}[htb]
    \centering
    \includegraphics[width=\textwidth]{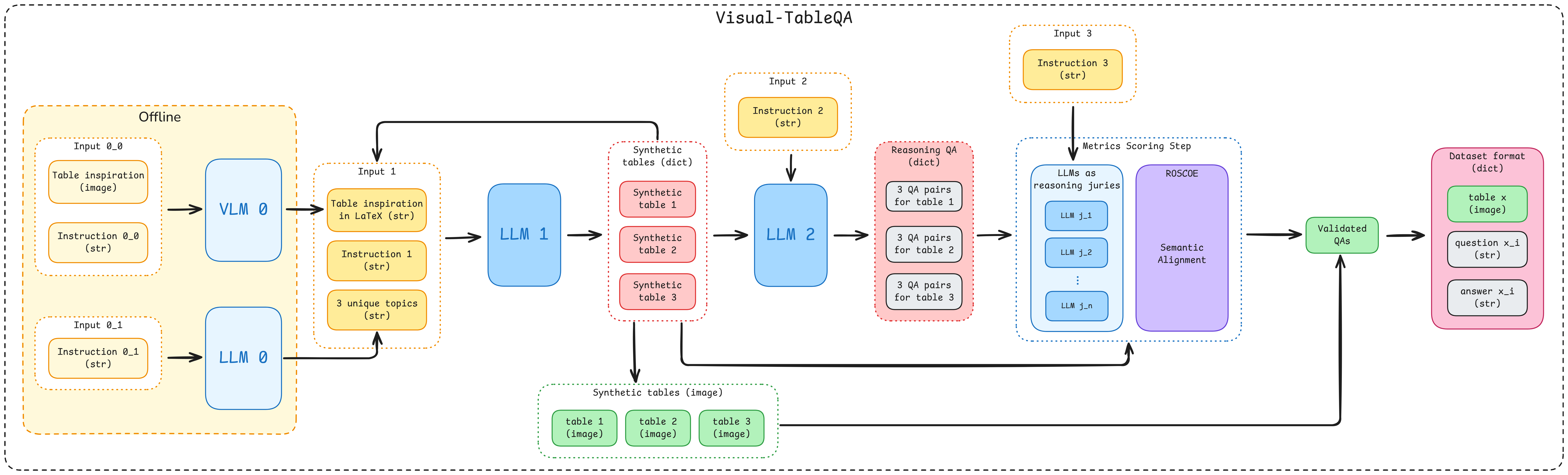}
    \caption{Overview of the full pipeline architecture of Visual-TableQA. A subset of initial table images is first converted to LaTeX using a visual language model (VLM-0). The resulting LaTeX code, along with topic prompts, is then passed to a language model (LLM-1) to generate new, diverse tables. These newly generated tables are used as inputs for further iterations of table generation. All generated tables are then submitted to a second language model (LLM-2), which produces corresponding question-answer pairs. Finally, the QA pairs are evaluated by a jury of high-performing LLMs, and their quality is assessed using the ROSCOE score.}
    \label{fig:overview}
\end{figure}

This section provides a detailed description of the generation pipeline. Figure~\ref{fig:overview} gives an overview of the whole process.

\noindent\textbf{\label{sec:seeds}Seed Tables and Topics Collection:}
The first step involves collecting a diverse set of table layouts to serve as inspiration for LLMs during the generation process. We explored various sources, including scientific journals, financial report databases, online newspapers, and table design galleries. Our search included both table and diagram images to introduce greater visual and structural complexity into the dataset. We selected 20 representative images (Figure~\ref{fig:layout-seeds}) and passed them to a visual language model, VLM-0 (\texttt{GPT-o3}~\citet{openai2025o3}), to generate accurate LaTeX representations. In parallel, we used LLM-0 (\texttt{GPT-4o})~\citet{openai2024gpt4o} to generate a list of 5,000 distinct topic prompts. These initial table samples and topics serve as the first layer of inspiration for subsequent LLM generations—though the pool of inspirations expands automatically, as detailed in Section~\ref{sec:layout-evolution}. For reproducibility, all resources are publicly available in our \href{https://github.com/AI-4-Everyone/Visual-TableQA}{GitHub repository}.

\noindent\textbf{Table Generation:}
For each iteration, we randomly select an LLM-1 from the models short-list presented in Table~\ref{tab:generation_accuracy}. The model receives one table sample from our pool and three topics randomly selected from the topic list, all delivered through a single instruction prompt. The output from LLM-1 is returned as a JSON file containing three newly generated LaTeX-formatted tables in plain text, each corresponding to one of the provided topics. We require that the generated tables be inspired by the input table but include substantial layout variations and, when appropriate, additional data to enhance complexity. The resulting LaTeX code is then compiled using standard LaTeX compilation stack (pdflatex + pdf2image), and cropped to produce high-resolution table images. A human reviewer then inspects the table and makes adjustments to the LaTeX code if necessary. The prompt used for generation are provided in Figure~\ref{fig:llm1-prompt}.

\noindent\textbf{\label{sec:layout-evolution}Evolving Layouts through Iterations:}
A subset of the generated tables is manually selected to enrich the pool of table inspirations. This feedback loop encourages the emergence of increasingly complex and diverse layouts by amplifying visual variations and enabling cross-model inspiration across different LLM-1s over successive iterations. This process is facilitated by the fact that LLMs differ in architecture and tend to focus on distinct structural and stylistic aspects of tables. As a result, combining inspirations across models leads to highly diversified and creative layout types. We refer to this phenomenon as \textbf{cross-model prompting (‘inspiration’)}.

\noindent\textbf{QA Generation:}
Next, for each generated table, we randomly select a model, denoted LLM-2, from the same list of models in Table~\ref{tab:generation_accuracy} to generate three QA pairs. The model receives the table in LaTeX format and is instructed to produce questions that require multi-step reasoning, pattern recognition, and symbolic interpretation. For instance, the sample in Figure~\ref{fig:qa_sample} illustrates how the questions extend beyond basic information extraction, requiring interpretative reasoning to identify patterns within the presented data. We do not fact-check the generated tables; as a result, some table content may be non-factual. While this is important to consider when using the dataset for training, it can be beneficial, as it encourages models to rely on reasoning rather than prior knowledge.

\subsection{\label{sec:quality}Quality Assurance}

To ensure the validity of the tables and QA pairs, a panel of independent LLMs—serving as a reasoning jury—evaluates each table and its associated QA pairs by providing binary correctness judgments. The evaluation is based on four criteria: \textit{(i)} the generated document is a valid table and is relevant to the given topic; \textit{(ii)} the table and any associated figures are coherent and meaningful; \textit{(iii)} the question is fully grounded in the table, requiring no external knowledge; and \textit{(iv)} the answer is completely supported by the table content. If any of these four criteria are not met, the corresponding table and its QA pairs are discarded. The LLM jury includes \texttt{Mistral-large}, \texttt{Deepseek-v3.1}, \texttt{Gemini-2.5-pro}, \texttt{GPT-4.1}, and \texttt{Deepcogito-v2}—models chosen for their strong reasoning abilities. Final acceptance is determined via majority vote across the jury. The prompt used is provided in Figure~\ref{fig:llm3-prompt}.

The next step involved computing the ROSCOE reasoning scores as introduced in~\citet{golovnevaroscoe}. These metrics assess the coherence, logical soundness, and contextual grounding of step-by-step generated rationales. The ROSCOE framework encompasses thirteen evaluation criteria, which we report in Table~\ref{tab:roscoe_sa} along with their corresponding values computed over our dataset. The results indicate near-perfect alignment with the expected directionality of each metric, supporting the overall quality of the generated reasoning chains.

\noindent\textbf{Test Set Construction and Human Evaluation:}
The dataset was divided into three subsets: training, validation, and testing. To prevent data leakage, all entries \{table, question, answer\} derived from a single table were assigned to the same subset. The testing set was also used for human evaluation. Two human annotators—each holding at least a Master’s degree and with prior experience in data annotation—were hired to evaluate the quality of 800 QA pairs. Each QA pair was assessed for validity and rated on a scale from 1 to 5. Overall, 92\% of the evaluated QA pairs received a rating of at least 4 stars from both annotators.

\section{Experiments}

\subsection{Benchmark Comparison}
\noindent\textbf{Evaluated Benchmarks and Model Selection:}
We evaluate a range of state-of-the-art reasoning VLMs on {Visual-TableQA} and compare their performance across three other benchmarks focused on table and chart-based visual question answering: {ChartQA}~\citet{masry2022chartqa}, {ReachQA}~\citet{hedistill}, and MATH-Vision~\citet{wang2024measuring}. Our model selection includes powerful proprietary models such as \texttt{GPT-4o}, \texttt{GPT-4o Mini}~\citet{openai2025_gpt4o_mini}, \texttt{Gemini 2.5 Flash}, \texttt{Gemini 2.5 Pro}, and \texttt{Claude 3.5 Sonnet}, as well as open-source models like \texttt{LLaMA 4 Maverick 17B-128E Instruct}, \texttt{Mistral Small 3.1 24B Instruct}~\citet{mistral2025_small3_1_24b_instruct}, \texttt{Qwen2.5-VL-32B-Instruct}~\citet{chen2024qwenvl}, \texttt{Qwen2.5-VL-7B-Instruct}~\citet{qwen2.5-VL}, \texttt{LLaVA-Next-Llama3-8B}~\citet{li2024llava}, \texttt{MiniCPM-V2.5-Llama3}~\citet{yao2024minicpm}, and \texttt{InternVL2-8B}~\citet{chen2024internvl}. Where performance metrics were available, we did not re-evaluate models on these datasets; instead, we report the results published in the original papers, official leaderboards, or model cards. For all other cases, we carefully fine-tuned and evaluated the models following the instructions provided in their respective official GitHub repositories.

\noindent\textbf{Evaluation Protocol:}
All models are evaluated on the test sets of the four selected datasets. Each model receives image-question pairs, formatted within a unified prompt that includes a system message tailored to elicit the model’s reasoning capabilities (Section~\ref{sec:jury-prompt}). For the Visual-TableQA dataset, we additionally construct a variant in which data is provided not as rendered images but in LaTeX code format. This textual-code version is referred to as Visual-TableQA-CIT.

For \texttt{LLaVA-Next-Llama3-8B}, \texttt{MiniCPM-V2.5-Llama3}, \texttt{InternVL2-8B}, and \texttt{Qwen2.5-VL-7B-Instruct}, we conducted two supervised fine-tuning (SFT) experiments: \textit{(i)} using the ReachQA training split (denoted as \texttt{Model\_Name + ReachQA}) and \textit{(ii)} using the Visual-TableQA training split (denoted as \texttt{Model\_Name + Visual-TableQA}). We applied Low-Rank Adapters (LoRA)~\citet{hulora} to all linear layers, following the SFT setup and hyperparameters described in the \citet{hedistill} \href{https://github.com/hewei2001/ReachQA/tree/main}{GitHub repository} when possible (Section~\ref{sec:hyperparam}) in order to make a fair comparison. The fine-tuning phase for all models was limited to one epoch to ensure consistency and reduce overfitting. Exceptionally, we adopted a custom two-phase LoRA fine-tuning strategy for \texttt{Qwen2.5-VL-7B-Instruct} (see Section~\ref{sec:hyperparam}), as this model was not included in the evaluation of~\cite{hedistill}, and to better accommodate the relatively small size of our dataset.

All models are allocated a maximum of 5,000 tokens during inference to accommodate extended chain-of-thought reasoning. Model responses are evaluated using the same jury of high-performing VLMs and majority-vote protocol as described in Section~\ref{sec:quality}. The jury confidence score, computed as the ratio of the highest vote count to the total jury size, averages above 0.87 for all models and all datasets. In addition, evaluations are run twice, to ensure reproducibility.

\subsection{Experimentation Results}
The average models accuracies are displayed in Table~\ref{tab:bench-results}. These results reveal that:

\noindent\textbf{Visual-TableQA Effectively Evaluates Visual Reasoning Capabilities:}
%\begin{wraptable}[26]{r}{0.53\linewidth}
\begin{figure}[htb]
\centering
\caption{Correlation of model rankings on Visual-TableQA with those on three established datasets—ChartQA (recognition-focused), ReachQA (balanced), and MATH-Vision (reasoning-focused)—using Spearman’s $\rho$ and Kendall’s $\tau$ metrics. Higher values indicate stronger alignment in model performance trends. Visual-TableQA shows strong correlation with ReachQA, suggesting it effectively balances both visual recognition and reasoning, while its weaker correlation with ChartQA and MATH-Vision highlights its unique position as a comprehensive visual reasoning benchmark.}
\includegraphics[width=0.6\linewidth]{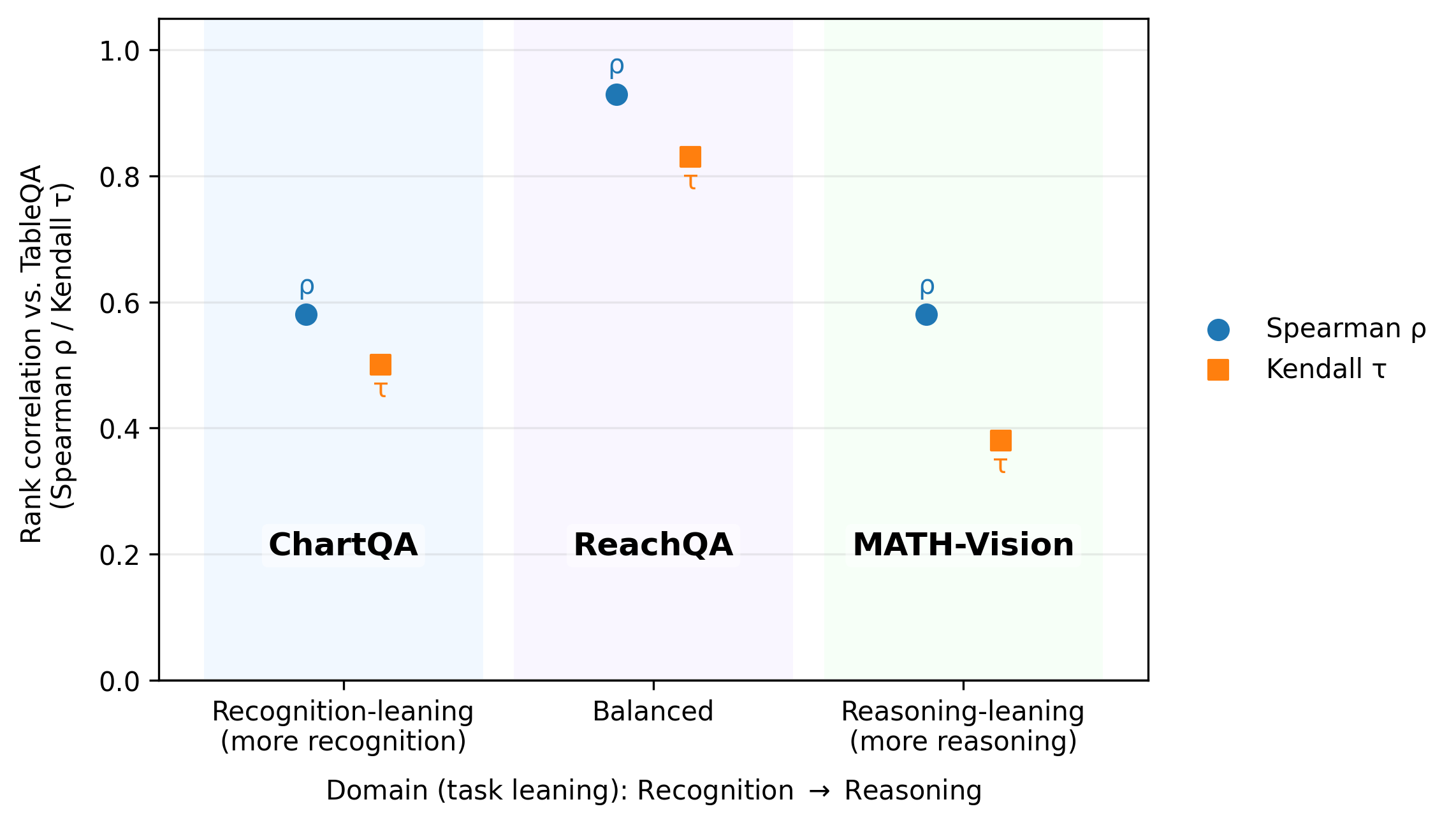}
\label{fig:rho_tau}
\end{figure}
%\end{wraptable}
Model performances on Visual-TableQA follow similar trends to those observed on real-world, human-annotated datasets such as ChartQA and MATH-Vision, suggesting that synthetic datasets can effectively evaluate reasoning capabilities.
A direct comparison between Visual-TableQA and its textual variant, Visual-TableQA-CIT, shows a notable performance gap: on average, models perform $+6.26\%$ better on Visual-TableQA-CIT. This highlights the added challenge posed by the image-based format in Visual-TableQA, demonstrating its effectiveness at testing visual reasoning over purely textual input.

To further validate Visual-TableQA as a reasoning benchmark, we compared model rankings across datasets. For each dataset, we extracted the models (except the fine-tuned ones) performance rankings and compared them to the rankings on Visual-TableQA using two correlation measures: \textit{(i)} Spearman’s $\rho$~\citet{lee1988thirteen}: Captures monotonic consistency in rankings (regardless of exact scores); \textit{(ii)} Kendall’s $\tau$~\citet{kendall1948rank}: Measures the fraction of concordant vs. discordant ranking pairs and is more robust to ties. Both metrics range from $-1$ to $1$, with values closer to $1$ indicating strong alignment in model rankings. To ensure fairness, we adjusted all scores computed with Relaxed Accuracy by subtracting 5\%, before comparison. The results are shown in Figure~\ref{fig:rho_tau}.

Each dataset varies in how much it emphasizes visual recognition versus reasoning:
\textit{(i)} ChartQA $\rightarrow$ Recognition-heavy, \textit{(ii)} ReachQA $\rightarrow$ Balanced, \textit{(iii)} MATH-Vision $\rightarrow$ Reasoning-heavy

Interestingly, Visual-TableQA rankings align most closely with ReachQA, but not with ChartQA or MATH-Vision individually. This suggests that Visual-TableQA does not favor models that excel solely at recognition or solely at reasoning. Instead, it rewards models capable of both—making it a comprehensive benchmark for evaluating all aspects of visual reasoning.

\noindent\textbf{Visual-TableQA Effectively Transfers to Other Benchmarks:}
To assess the transferability of {Visual-TableQA}, we investigated how fine-tuning on Visual-TableQA impacts performance across other benchmarks. As shown in Table~\ref{tab:bench-results}, supervision from Visual-TableQA led to significant generalization beyond its native domain. Notably, it improved the accuracy of \texttt{Qwen2.5-VL-7B-Instruct} on \textit{ReachQA} from 49.23\% to 60.95\%, and on \textit{MATH-Vision} from 25.10\% to 49.77\%, despite these datasets not being explicitly table-focused. This finding is further supported by Table~\ref{tab:reco_vs_reas}, which reports similar gains in generalization across three additional models: \texttt{LLaVA-Next-Llama3-8B}, \texttt{MiniCPM-V2.5-Llama3}, and \texttt{InternVL2-8B}.

However, this transferability is not reciprocal. Fine-tuning \texttt{Qwen2.5-VL-7B-Instruct} on \textit{ReachQA} alone yields only modest in-domain gains (49.23\% $\rightarrow$ 55.75\%) and leads to reduced performance on both \textit{ChartQA} and \textit{Visual-TableQA}. This suggests that {Visual-TableQA} provides a more generalizable reasoning signal—rooted in layout understanding, symbolic interpretation, and multi-step reasoning—compared to standard benchmarks.

\noindent\textbf{Proprietary Models Outperform Open-Source Models on Average:}
\texttt{Claude 3.5 Sonnet} achieves the highest performance across nearly all benchmarks. However, fine-tuning on {Visual-TableQA} substantially narrows the gap between proprietary and open-source models. Notably, the performance of \texttt{Qwen2.5-VL-7B-Instruct} increases significantly across all evaluated benchmarks—surpassing several state-of-the-art proprietary models, including \texttt{GPT-4o}, \texttt{GPT-4o-mini}, and \texttt{Gemini 2.5 Pro}.

\section{Discussion}

\subsection{Visual-TableQA vs ReachQA}
\begin{table}[htb]
    \centering
    \scriptsize
    \caption{Performance of fine-tuned models on the two splits of the ReachQA test set: \textit{Recognition} (Reco) and \textit{Reasoning} (Reas), each consisting of exactly 1,000 samples. Best performances per model category are in \textbf{bold}. The values in {\color{blue}blue} are from our own evaluation using the LLM jury, while the remaining values are taken from ~\citet{hedistill}.}
    \begin{tabularx}{\textwidth}{X p{1cm} p{1cm}@{\hspace{11em}} X p{1cm} p{1cm}}
    \toprule
     \textbf{Model} & \textbf{Reco} & \textbf{Reas} & \textbf{Model} & \textbf{Reco} & \textbf{Reas}\\
     \midrule
     LLaVA-Next-Llama3-8B & 17.9 & 6.5 & InternVL2-8B & 33.7 & 16.2\\
     + \textbf{ReachQA} & \textbf{29.6} & 11.1 & + \textbf{ReachQA} & \textbf{49.8} & 21.3\\
     + \textbf{Visual-TableQA} & \textcolor{blue}{28.4} & \textbf{\textcolor{blue}{20.2}} & + \textbf{Visual-TableQA} & \textcolor{blue}{45.6} & \textbf{\textcolor{blue}{34.5}}\\
     \midrule
     MiniCPM-V2.5-Llama3 & 25.3 & 10.3 & Qwen2.5-VL-7B-Instruct & \textcolor{blue}{66.20} & \textcolor{blue}{33.10} \\ 
     + \textbf{ReachQA} & 35.10 & 11 & + \textbf{ReachQA} & \textcolor{blue}{69.6} & \textcolor{blue}{40.30} \\ 
     + \textbf{Visual-TableQA} & \textbf{\textcolor{blue}{36.20}} & \textbf{\textcolor{blue}{31.50}} & + \textbf{Visual-TableQA} & \textbf{\textcolor{blue}{70.3}} & \textbf{\textcolor{blue}{50.6}}\\
     \addlinespace[0.15em]
     \midrule
     \multicolumn{6}{c}{\textbf{Average gains}}\\
     \midrule
     + \textbf{ReachQA} & \textbf{\textcolor{nicegreen}{$+10.25$}} & \textcolor{nicegreen}{$+4.4$}\\
     + \textbf{Visual-TableQA} & \textcolor{nicegreen}{$+9.35$} & \textcolor{nicegreen}{$+17.68$}\\
     \bottomrule
    \end{tabularx}
    \label{tab:reco_vs_reas}
\end{table}

The ReachQA dataset is divided into two equally sized subsets: \textit{Recognition}, which tests a model’s ability to extract relevant information from charts, and \textit{Reasoning}, which evaluates a model’s capacity to understand complex and abstract data structures. Table~\ref{tab:reco_vs_reas} reports the performance gains of multiple fine-tuned models on these two tasks.

On average, models fine-tuned on ReachQA exhibit an accuracy improvement of $+10.25$ points on the \textit{Recognition} task and $+4.4$ points on the \textit{Reasoning} task. In comparison, models fine-tuned on {Visual-TableQA} show an average gain of $+9.35$ on \textit{Recognition}—a comparable result—but a significantly larger gain of $+17.68$ on \textit{Reasoning}.

This stark contrast in reasoning performance can be attributed to the presence of high-quality rationales in Visual-TableQA annotations, along with the inclusion of more complex and diverse visual structures. In other words, despite being roughly three times smaller than ReachQA in terms of sample count, Visual-TableQA places a stronger emphasis on qualitative richness over quantity. As a result, it appears to enable more effective knowledge distillation, particularly for tasks requiring symbolic interpretation and multi-step reasoning.

\subsection{Visual-TableQA’s Advantages Compared to Other Datasets}

Table~\ref{tab:dataset-comparison} shows that only a few table-focused QA datasets—namely TAT-DQA, Table-VQA, and TableVQA-Bench—represent tables as rendered images. \textbf{Visual-TableQA} surpasses these by offering richer layout diversity, broader topic coverage, systematic visual complexity, and high-quality rationales. These attributes make it particularly effective for training models with transferable reasoning skills. Supporting this, models fine-tuned solely on \textbf{Visual-TableQA}—such as \texttt{LLaVA-Next-Llama3-8B}—demonstrated significant gains on external benchmarks (Table-VQA and TableVQA-Bench), as seen in Table~\ref{tab:tableqa_vs_tablevqa}.

Interestingly, \texttt{Qwen2.5-VL-7B-Instruct} did not follow the same performance trend: it showed degradation on tasks such as \textit{VTabFact} (Yes/No fact verification), \textit{VWTQ} (Wikipedia table retrieval), and \textit{VWTQ-Syn} (synthetic variants). To understand this, we manually analyzed its errors before and after fine-tuning on \textit{VTabFact}, categorizing them into eight types: \textit{partial data extraction, hallucination, incoherence, misunderstanding, reasoning errors, evaluation mistakes, dataset ambiguity}, and \textit{annotation flaws}. Results (Figure~\ref{fig:vtab-error-taxonomy}) show that while the total number of errors slightly increased post-finetuning, most now fall into the \textit{incoherence} class, with all other error types significantly reduced. This suggests a sharpening of reasoning patterns but also highlights a need for future work targeting specific error types through synthetic supervision. Further details are provided in Section~\ref{sec:errors}.

Beyond transferability and diversity, a key advantage of Visual-TableQA lies in its modularity and scalability as explained in Section~\ref{sec:scalability} . 

\begin{table}[htb]
    \centering
    \scriptsize
    \caption{Performance of fine-tuned models on Table-VQA test set and the four splits of the TableVQA-Bench dataset: \textit{FinTabNetQA} (finance-related tables), \textit{VTabFact} (table-based fact verification with Yes/No questions), \textit{VWTQ} (information retrieval from Wikipedia tables), and \textit{VWTQ-Syn} (synthetic visual variants of VWTQ). Best performances per model variants are shown in \textbf{bold}. Values in {\color{blue}blue} are from our own evaluation using the \texttt{DeepSeek-Prover-v2}, while remaining values are reported from ~\citet{fu2025refocus}.
}
    \renewcommand{\arraystretch}{1.3}
    \begin{tabularx}{0.9\textwidth}{>{\raggedright\arraybackslash}X *{4}{>{\arraybackslash}p{1.3cm}} >{\centering\arraybackslash}p{1.5cm} }
    \toprule
    \multirow{2}{=}{\textbf{Model}} & \multicolumn{4}{c}{\centering\textbf{TableVQA-Bench}} & \multirow{2}{=}{\centering\textbf{Table-VQA}}\\
    \cmidrule(lr){2-5}% \cmidrule(lr){3-3} \cmidrule(lr){4-4} \cmidrule(lr){5-5}
    & \textbf{FinTabNetQA} & \textbf{VTabFact} & \textbf{VWTQ} & \textbf{VWTQ-Syn} & \\
    \midrule
    GPT-4o & \textcolor{blue}{96.8} & \textcolor{blue}{78.0} & \textcolor{blue}{72.8} & \textcolor{blue}{82.4} & \textcolor{blue}{--}\\
    LLaVA-Next-34B & \textcolor{blue}{--} & 71.2 & 36.4 & 38.0 & \textcolor{blue}{--}\\
    \hdashline
    LLaVA-Next-Llama3-8B & \textcolor{blue}{52.4} & \textcolor{blue}{37.2} & \textcolor{blue}{21.5} & \textcolor{blue}{24.8} & \textcolor{blue}{25.84}\\
    + \textbf{Visual-TableQA}  & \textcolor{blue}{\textbf{56.8}} & \textcolor{blue}{\textbf{52.0}} & \textcolor{blue}{\textbf{33.2}} & \textcolor{blue}{\textbf{33.6}} & \textcolor{blue}{\textbf{28.89}}\\
    \hdashline
    Qwen2.5-VL-7B-Instruct & \textcolor{blue}{96.4} & \textcolor{blue}{\textbf{82.0}} & \textcolor{blue}{\textbf{68.53}} & \textcolor{blue}{\textbf{74.0}} & \textcolor{blue}{\textbf{79.03}}\\
    + \textbf{Visual-TableQA} & \textbf{\textcolor{blue}{97.2}} & \textcolor{blue}{70.6} & \textcolor{blue}{61.5} & \textcolor{blue}{69.6} & \textcolor{blue}{75.23}\\
    \bottomrule
\end{tabularx}
    \label{tab:tableqa_vs_tablevqa}
\end{table}

\subsection{\label{sec:scalability}Scalability of the Pipeline and Its Benefits for Knowledge Distillation}

This modular pipeline supports scalable generation with a clean separation of concerns—table structure synthesis, QA creation, and validation—making each component independently reusable and upgradable. By automating the entire process from table generation to jury-based quality control, Visual-TableQA provides a cost-efficient and high-quality benchmark for advancing multimodal reasoning over complex visual inputs. A central component of our pipeline is the mechanism of \textbf{cross-model inspiration}~\ref{sec:layout-evolution}, a collaborative prompting strategy. In this process, stronger models generate layout ``seeds'' that guide weaker models in synthesizing structurally diverse tables, fostering novel visual configurations through iterative transfer. The same principle extends to question–answer generation: models are prompted with both layout and topical cues—often proposed by stronger models—to create new QA pairs. This enables weaker models to contribute meaningfully to the dataset by expanding the range of questions and reasoning patterns. Through this dual-inspiration process, the pipeline cultivates a collaborative multi-model co-creation space, where models of varying capabilities distill collective knowledge not through imitation, but through generative
inspiration, while maintaining data quality. In this regard, \textbf{Visual-TableQA} distinguishes itself from other synthetic datasets~\citet{aboutalebi2024magid, wang2024graph, li2025smir, hedistill}.

\section{Conclusion}

In this work, we introduced {Visual-TableQA}, a large-scale, open-domain, multimodal dataset designed to rigorously evaluate visual reasoning capabilities over complex table images. Building on the principles of Code-as-Intermediary Translation (CIT), we developed a fully automated, modular pipeline for generating LaTeX-rendered tables, reasoning-intensive question–answer pairs, and high-quality rationales—all verified by a jury of strong LLMs. Despite being cost-efficient (generated for under \$100), {Visual-TableQA} offers unprecedented diversity in table structures, visual features, and reasoning depth. We showed that {Visual-TableQA} not only challenges existing visual language models (VLMs) but also serves as an effective training signal for improving reasoning performance. Fine-tuning on Visual-TableQA led to substantial gains across multiple benchmarks—both table-centric and general-purpose—including ReachQA and MATH-Vision, demonstrating the dataset's capacity to bridge the performance gap between open-source and proprietary models. 

\bibliography{iclr2026_conference}
\bibliographystyle{iclr2026_conference}

\appendix
\section*{Appendix}
\section{\label{related} Extended Related Works}
The vast majority of table-based QA datasets—such as HybridQA~\citet{chen-etal-2020-hybridqa}, WikiTableQuestions~\citet{pasupat-liang-2015-compositional}, WikiSQL~\citet{zhongSeq2SQL2017}, and AIT-QA~\citet{katsis-etal-2022-ait}—represent tables in textual format rather than as rendered images, thereby bypassing the challenges associated with visual layout interpretation. In contrast, our work focuses exclusively on multimodal datasets—those that contain both textual and visual (image-based) information. These can generally be grouped into two main categories: real-world datasets, collected from authentic documents, and synthetic datasets, generated using automated tools. Real-world multimodal QA datasets that emphasize tables—such as TAT-DQA~\citet{zhu2022towards}, and TableVQA-Bench~\citet{kim2024tablevqa}—tend to be highly domain-specific, limiting diversity in both table layouts and question types. For example, TAT-DQA~\citet{zhu2022towards} combines tabular and textual data from financial reports and, while it introduces hybrid contexts for realistic reasoning, its questions rely heavily on reading textual input rather than interpreting visual structure. Similarly, TableVQA-Bench~\citet{kim2024tablevqa} consists of 83\% real-world tables (1,250 out of 1,500), primarily sourced from task-specific datasets such as WikiTableQuestions (information retrieval), TabFact~\citet{Chen2020TabFact:} (fact verification), and FinTabNet~\citet{zheng2021global} (financial data extraction). Due to its relatively small size and the specialized nature of its subsets, the dataset exhibits limited visual diversity. This limitation also extends to the remaining 16\% synthetic tables, whose visual variation is restricted to basic formatting attributes such as background color, border size, font size, and style. More recently, ChartQA~\citet{masry2022chartqa} and DocVQA~\citet{mathew2020docvqa} have introduced large open-domain datasets for visual question answering. ChartQA focuses on reasoning over charts and plots; however, its tasks primarily involve shallow reasoning and do not reflect the structural complexity and layout diversity found in real-world tables. In contrast, DocVQA offers greater diversity in document layouts and structures, but lacks significant visual challenge—recent VLMs, including relatively lightweight models like \texttt{Qwen2.5-VL-7B-Instruct}~\citet{qwen2.5-VL}, achieve over 94\% accuracy on this benchmark.

Among synthetic datasets, \textit{MultiModalQA}~\citet{talmor2021multimodalqa} stands out as the only open-domain resource focused on tables. It combines real-world figures, diagrams, and text passages sourced from Wikipedia, with QA pairs crafted to assess both reasoning and visual comprehension. Although it incorporates real content, the dataset is considered synthetic due to the way it links independent modalities and generates QA pairs through formalized templates. However, this approach results in limited diversity, as the questions are derived from a finite set of templates. In contrast, \textit{Table-VQA}~\citet{AgDeTQA} is a fully synthetic dataset generated using state-of-the-art LLMs. Nonetheless, it lacks visual diversity and complexity—its tables follow similar formats and are predominantly centered around technical domains such as statistics, physics, and algorithms, all of which are heavily numerical in nature.

Most of real-world datasets rely heavily on manual labeling, data collection, and preprocessing—factors that significantly constrain their scalability. Recently, ReachQA~\citet{hedistill} introduced a more scalable and innovative approach through its Code-as-Intermediary Translation (CIT) pipeline. This method generates synthetic charts and reasoning questions by leveraging textual intermediaries such as Python code, demonstrating that advanced reasoning capabilities of large language models (LLMs) can be effectively transferred to visual models. While ReachQA successfully addresses both scalability and reasoning complexity, its approach is tailored to chart-based visualizations and does not extend to structured tabular data.

To summarize, existing table-based benchmarks consistently fall short in one or more key areas: visual diversity, reasoning depth, or scalability. Notably, aside from MultiModalQA, there is no open-domain dataset designed to evaluate model performance on rendered table images, despite their prevalence in real-world settings such as reports, academic papers, and spreadsheets. In this work, we introduce \textbf{Visual-TableQA}, a multimodal open-domain synthetic dataset specifically created to assess reasoning capabilities over table images using LLMs

\section{Considerations and Limitations}
The main limitations of this work relate to the use of Code-as-Intermediary Translation (CIT)~\citet{hedistill} and the assessment of data quality. While we adopted LaTeX as an intermediate representation for tables, its expressiveness is limited when handling more complex or visually rich images. Developing a robust, bidirectional image-to-text encoding system remains an open and promising area for future research. In terms of data quality evaluation, although automatic metrics such as ROSCOE provide useful insights, they are not yet as reliable as human judgment. As a result, human annotators continue to play a critical role in ensuring high-quality data, especially when scaling synthetic datasets for reasoning tasks.

In addition, we also observed that certain models, such as \texttt{Qwen2.5-VL-7B-Instruct}, did not consistently benefit from Visual-TableQA supervision across all downstream tasks, highlighting a potential limitation in generalization that warrants further investigation.

\section{TableQA Layout and Topic Diversity}

\begin{figure}[htb]
    \centering
    \includegraphics[width=\textwidth]{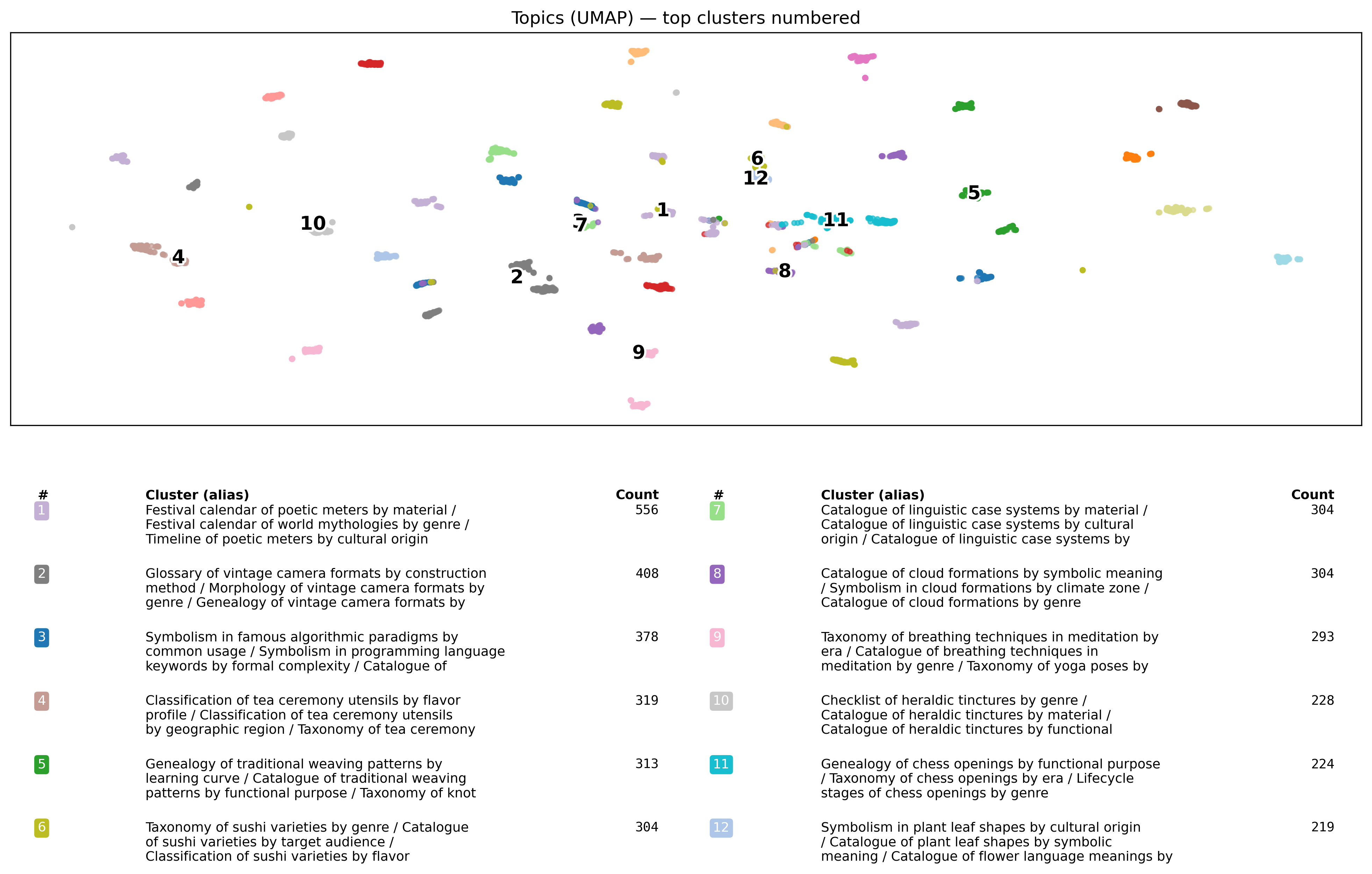}
    \caption{2D projection of the 5,000 topics using UMAP and K-Means clustering. Each color denotes a semantic cluster. Representative topics are listed for the 12 largest clusters to illustrate diversity.}
    \label{fig:topics-map}
\end{figure}

As described in Section~\ref{sec:seeds}, we sampled 5,000 distinct topics using \texttt{GPT-4o} to serve as inspirations for table generation. To better illustrate topic diversity, we grouped these topics into 20 semantic clusters using the K-Means algorithm. Figure~\ref{fig:topics-map} displays a 2D projection of these clusters, where each color represents a distinct semantic group. For the 12 largest clusters, we highlight representative topics to give a sense of their thematic content.

\begin{figure}[htb]
    \centering
    \includegraphics[width=0.5\linewidth]{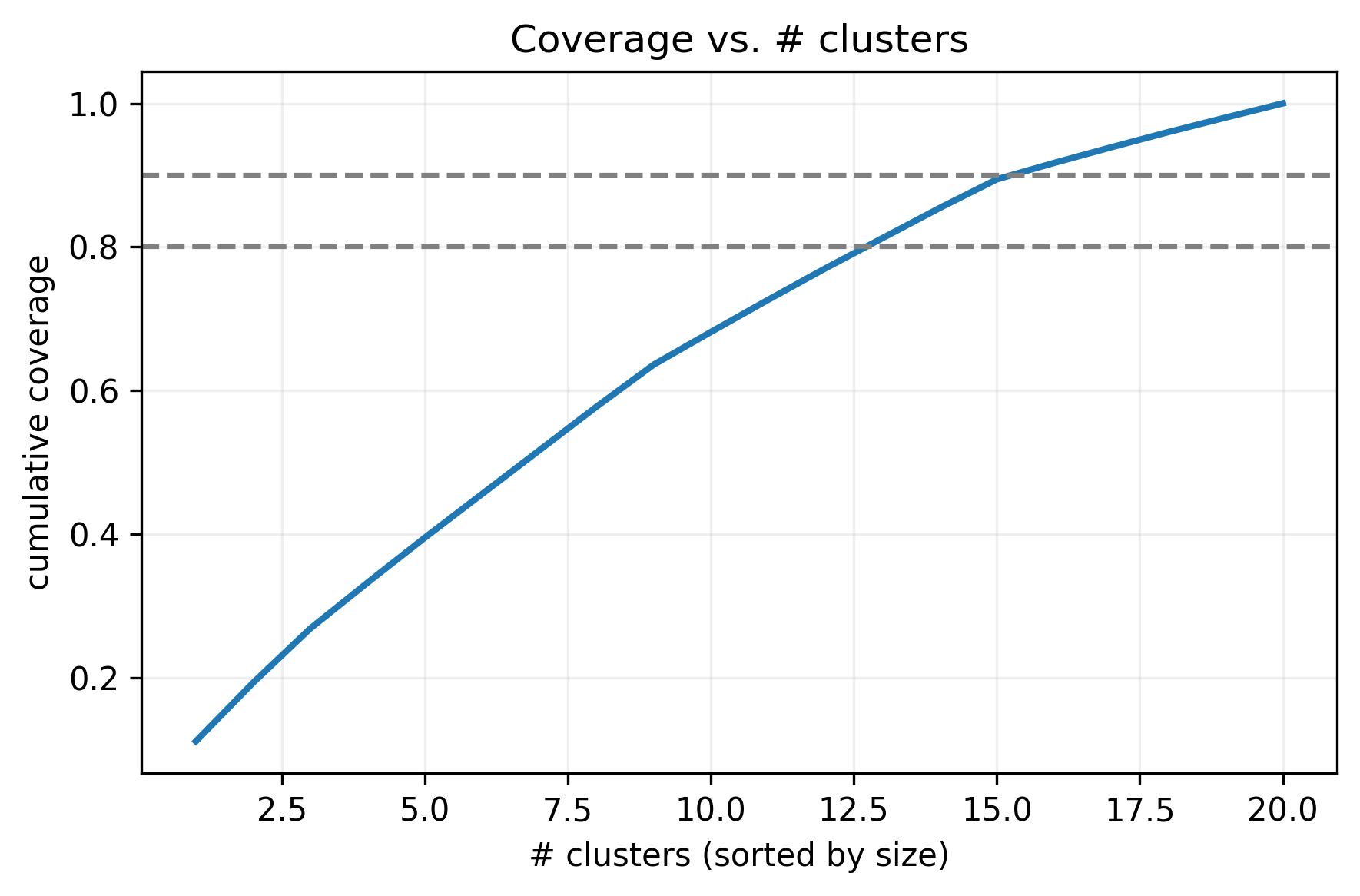}
    \caption{Cumulative topic coverage as clusters are added by descending size. The uniform slope indicates an even distribution of topics across clusters.}
    \label{fig:topics-coverage}
\end{figure}

In addition, Figure~\ref{fig:topics-coverage} shows the cumulative percentage of topics covered as clusters are added in descending order of size. The smooth progression of the curve indicates that the clusters are relatively uniform in size, confirming a balanced distribution of topic diversity throughout the dataset.

\begin{figure}[htb]
    \centering
    % First image
    \begin{subfigure}{0.43\textwidth}
        \centering
        \includegraphics[width=\linewidth]{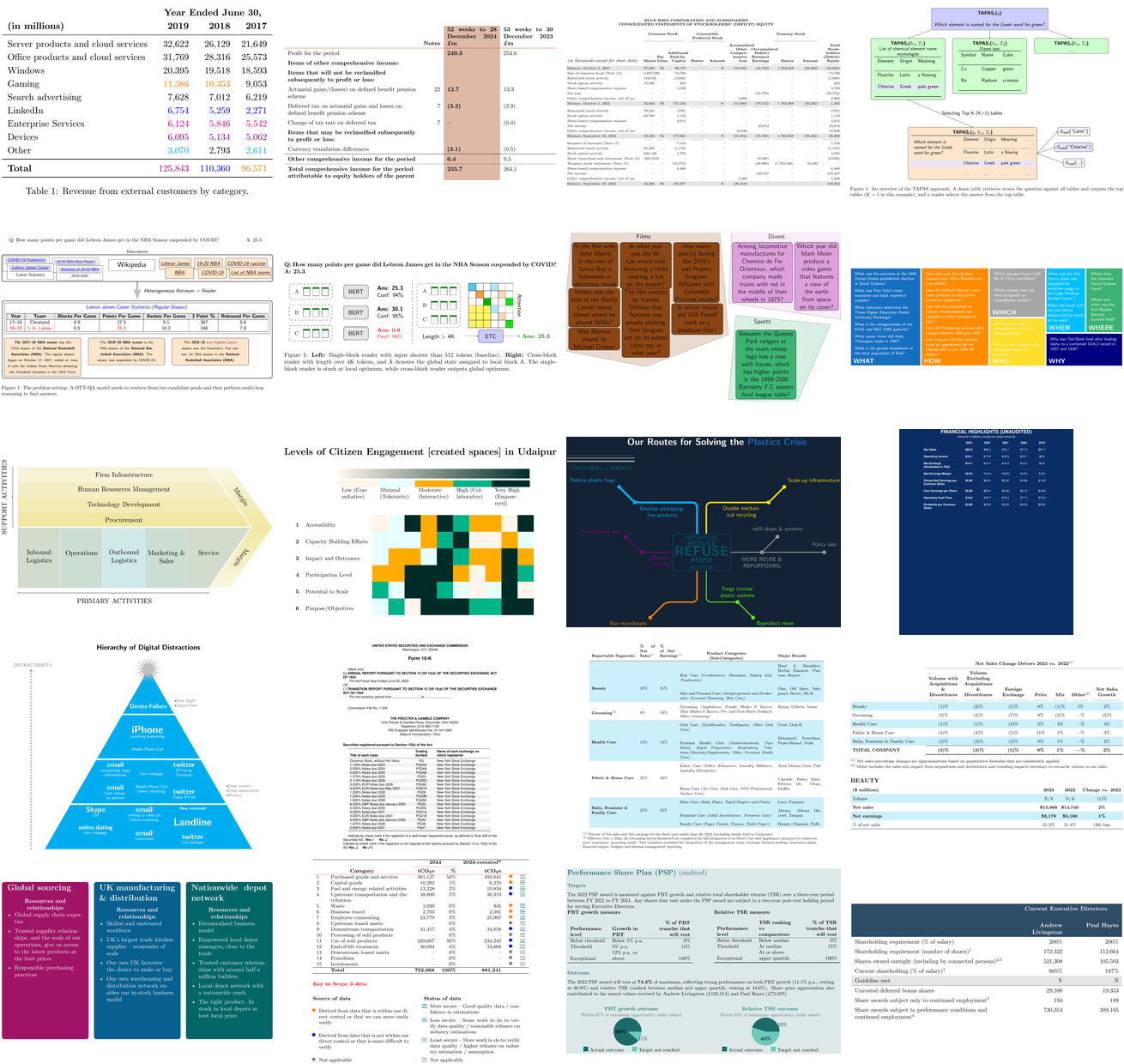}
        \caption{Layout seeds used for the initial table generation.}
        \label{fig:layout-seeds}
    \end{subfigure}
    \hfill
    % Second image
    \begin{subfigure}{0.55\textwidth}
        \centering
        \includegraphics[width=\linewidth]{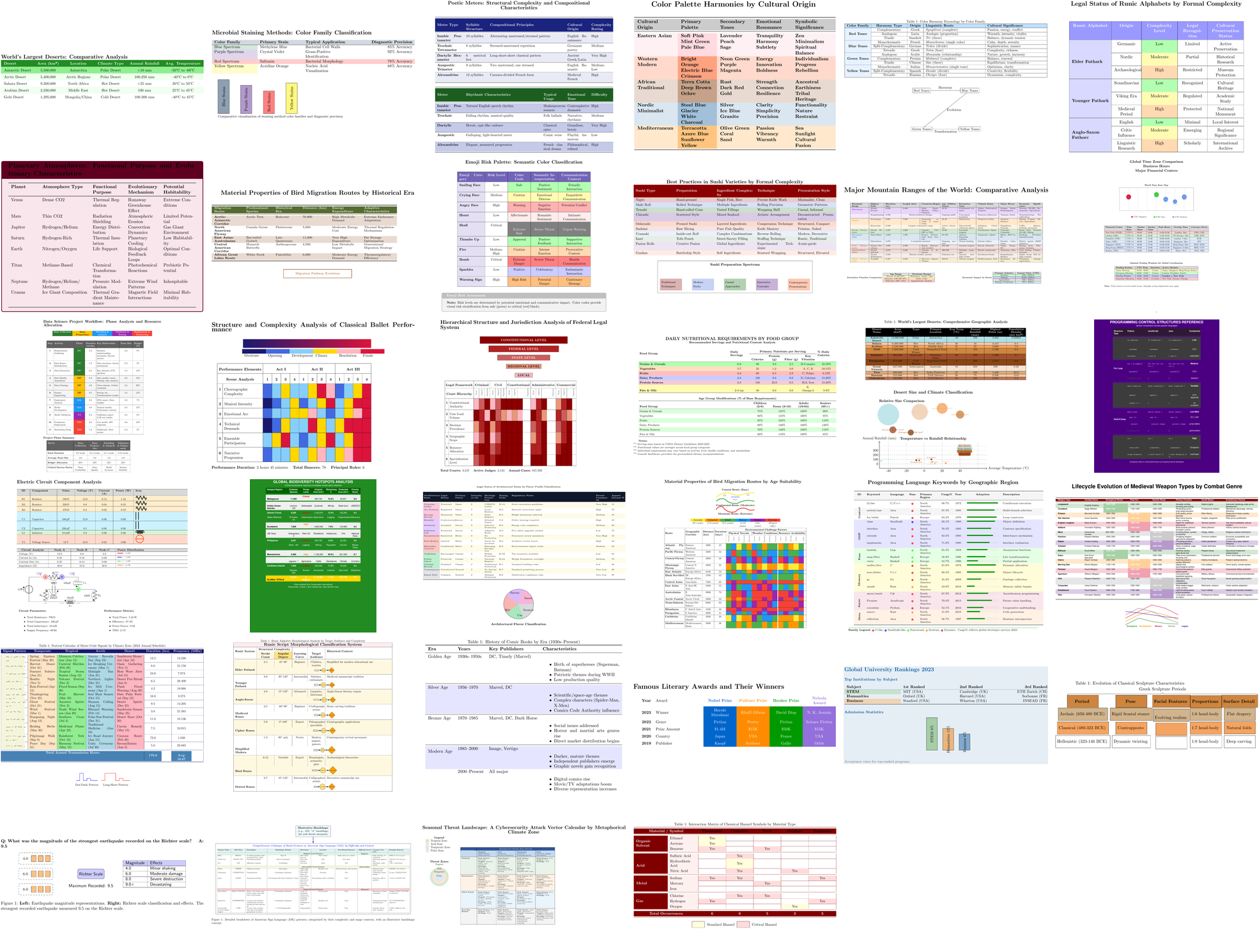}
        \caption{Sample of diverse layouts generated by our pipeline.}
        \label{fig:layout-samples}
    \end{subfigure}
    \caption{Visual diversity of table layouts. Left: seed layouts used during the first iteration of table generation. Right: layouts generated through cross-model inspiration and iterative refinement.}
    \label{fig:layouts}
\end{figure}

To illustrate the diversity of table layouts produced by our pipeline, Figure~\ref{fig:layouts} displays a side-by-side comparison between the initial seed tables used in the first generation iteration and a sample of layouts generated in subsequent steps. The wide range of structures highlights the pipeline’s capacity to create rich and varied visual designs from limited starting templates.

\section{Table Generation Settings}

The first stage of the generation pipeline involves \textbf{LLM-1}, which is responsible for producing new tables based on given inspirations. Specifically, it receives one LaTeX-formatted table as a \textit{layout inspiration} and three distinct \textit{topic inspirations}. Based on these, it generates three new tables, each aligned with one of the provided topics while drawing structural influence from the layout example. The full prompt used to guide LLM-1 during this step is shown in Figure~\ref{fig:llm1-prompt}.

The second stage of the generation pipeline involves \textbf{LLM-2}, which is responsible for producing question–answer (QA) pairs based on a single LaTeX table. The full prompt used to guide LLM-2 is shown in Figure~\ref{fig:llm2-prompt}.

To encourage creativity while maintaining factual accuracy, the temperature parameter for each model during this phase is set to 0.7.

\begin{figure}[htb]
\centering
\begin{minipage}{\textwidth}
\begin{lstlisting}
You are an expert in generating synthetic datasets composed of LaTeX-formatted tables, optionally accompanied by illustrative diagrams. Your task is to produce structured content suitable for data-centric documents, ensuring each table (and diagram, if included) is clear, well-organized, and visually informative.
Your final output should start with ```json and end with ``` as plain text, not just formatting. Like this:

```json
{
  "table_1": "BEGIN_LATEX
<LaTeX code for table 1 (with/without diagram) here>
END_LATEX",
  "table_2": "BEGIN_LATEX
<LaTeX code for table 2 (with/without diagram) here>
END_LATEX",
  "table_3": "BEGIN_LATEX
<LaTeX code for table 3 (with/without diagram) here>
END_LATEX"
}
```

Requirements:
    The tables and diagrams will be used to generate reasoning questions. Therefore:

        - If topic inspirations are supplied, ensure every generated table aligns with those topics.
        - Each LaTeX output must primarily consist of a table. Include a diagram only if it meaningfully complements the table; avoid adding one unnecessarily. Do not generate diagrams alone. If a diagram is empty or non necessary DON'T INCLUDE it.
        - Keep any diagram minimal—smaller than the table, chart-free, and purely illustrative—serving only to reinforce the table’s content without adding new information.
        - Each table and their diagram must contain realistic, domain-relevant content. They must be self-contained, include a clear descriptive title and not rely on external data to compile.
        - The type of information presented should be diverse—such as numerical data or qualitative. The variety and richness of visual elements is essential to the overall quality of the table and their diagram. Table quality should also come with a large number of rows and columns.
        - Table and diagram layouts should be creatively designed—taking inspiration from reference example (when provided) but incorporating meaningful variations such as colors, multi-row or multi-column cells, custom formatting adjustments, or any other visual enhancement that promotes structural diversity.     
        - Table layouts should be at least as complex as the example provided, don't try to simplify (diagrams are not mandatory). Table complexity should also come with a large number of rows and columns.
        - Do NOT escape any characters in the LaTeX code. The LaTeX must be written as plain text, exactly as it would appear in a .tex file, with real line breaks and single backslashes (\), not JSON-escaped.
        - All LaTeX tables and diagrams must be constrained to fit entirely within the printable area of a standard A4 page when compiled to PDF, without overflowing horizontally or vertically. Use appropriate formatting techniques such as adjusting column widths, reducing font size, or enabling landscape mode if necessary but NEVER rotation.
        - Make sure each LaTeX table and diagram includes all required \usepackage declarations and is enclosed within a complete, compilable LaTeX document structure, including the appropriate preamble and \begin{document}...\end{document} block.  
        - Make sure each LaTeX codes start and end with BEGIN_LATEX and END_LATEX, respectively.
        - Make sure to wrapp your final answer with ```json at the beginning and ``` at the end.
\end{lstlisting}
\end{minipage}
\caption{LLM prompt used for table generation.}
\label{fig:llm1-prompt}
\end{figure}

\begin{figure}[htb]
\centering
\begin{minipage}{\textwidth}
\begin{lstlisting}
You are an expert in generating question–answer pairs from LaTeX-formatted. Your task is to create a structured dataset consisting of visually challenging, reasoning-based questions and their corresponding answers derived from a given LaTeX formatted table with optional diagram.

Input:

You will be provided with a sample LaTeX table as context. Based on this table or diagram, your goal is to generate a JSON object with the following structure:

    questions: A python list of 3 challenging questions that require reasoning and analysis based ONLY on the data presented in the table and the optional diagram. The questions must be answerable using ONLY the information in the table or diagram(no extra knowledge).
    answers: A python list of 3 detailed answers to the 3 questions, including a clear chain of thought explaining the reasoning process.

Requirements:

    All questions must be relevant to the table's context and designed to test deeper understanding or inference.
    When possible, all questions should make full use of the visual or structural elements of the table or diagram (such as rows, columns, headers, colors, patterns, diagrams etc.) while maintaining clear relevance to the table’s content.
    Questions must be clear and answarable with an objective methodology, no subjective question.
    All entries (both questions and answers) should be returned as lists of string values.
    The global result should be a single JSON object wrapped in a markdown code block using ```json at the beginning and ``` at the end, and containing all two key-value pairs.
    This means your output should start with ```json and end with ``` as plain text, not just formatting.
\end{lstlisting}
\end{minipage}
\caption{LLM prompt used for QA generation.}
\label{fig:llm2-prompt}
\end{figure}

In Table~\ref{tab:qa_accuracy}, we present the average validity rates of the generated QA pairs for each model involved in the QA generation phase.

\begin{table}[htb]
\centering   
\scriptsize
\caption{Average QA pair validity across different QA generation models. Accuracies are computed from a sample of at least 500 QA pairs per model.}
\begin{tabularx}{0.55\columnwidth}{X >{\centering\arraybackslash}p{1.5cm}}
\toprule
\textbf{Model} & \textbf{Acc. (\%)}\\
\midrule
Llama 4 Maverick 17B-128E Instruct & 88 \\
Gemini 2.0 Flash & 89.1 \\
Gemini 2.5 Flash & 93.1 \\
Gemini 2.5 Pro & 89.3 \\
GPT-4.1 & 90.4 \\
Qwen3-30B-A3B & 76.6 \\
Qwen3-QwQ-32B & 92.6 \\
DeepSeek-R1-Distill-Qwen-32B & 73.4 \\
DeepSeek-R1T-Chimera & 89.4 \\
Claude Sonnet 4 & 91 \\
Claude 3.5 Haiku & 90.2 \\
Grok 3 Beta & 89.4 \\
Reka Flash 3 & 79 \\
\bottomrule
\end{tabularx}
\label{tab:qa_accuracy}
\end{table}

\section{Common Anomalies in Generated Tables}

As summarized in Table~\ref{tab:generation_accuracy}, several tables generated by different LLMs required post-processing adjustments to ensure visual clarity, LaTeX correctness, and topic alignment. Below, we categorize the most common types of anomalies observed during generation:

\subsection*{Layout and Formatting Issues}
\begin{itemize}
    \item \textbf{Table overflow:} Tables exceeding page margins due to improper column widths or missing column dimensioning.
    \item \textbf{Text overflow:} Cell content spilling outside the cell boundary, especially in narrow columns or with long strings.
    \item \textbf{Invisible content:} Multirow cells with background colors that obscure cell text (e.g., white text on white background).
    \item \textbf{Improper horizontal lines:} \texttt{\textbackslash midrule} or \texttt{\textbackslash hline} splitting across multirow cells, breaking visual coherence.
\end{itemize}

\subsection*{Content Relevance and Correctness}
\begin{itemize}
    \item \textbf{Empty or irrelevant tables:} Tables with placeholder content or unrelated to the assigned topic.
    \item \textbf{Incorrect topic alignment:} Generated tables that do not match the intended topic inspiration.
    \item \textbf{Duplicate outputs:} All three tables generated for a prompt are identical or nearly identical in structure/content.
    \item \textbf{Missing external resources:} References to images or figures not included or available in the output.
    \item \textbf{Incorrect LaTeX syntax:} Math symbols placed outside of math environments, leading to compilation errors.
\end{itemize}

\subsection*{Diagram-Specific Issues}
\begin{itemize}
    \item \textbf{Missing tables:} Some generations return only a diagram without an accompanying table.
    \item \textbf{Node placement errors:} Overlapping or misaligned nodes in TikZ diagrams.
    \item \textbf{Arrow misplacement:} Arrows that do not connect to correct nodes or that overlap diagram elements improperly.
    \item \textbf{Legend/title confusion:} Titles or legends positioned incorrectly or detached from the relevant diagram elements.
    \item \textbf{Visual inconsistencies:} General drawing flaws, such as missing anchors, inconsistent line styles, or unintended overlaps.
\end{itemize}

These issues highlight the need for a validation loop in the TableQA pipeline and justify the inclusion of human verification stages to ensure dataset quality.

\section{ROSCOE Metric Scores}
In addition to the LLM jury validation of the Visual-TableQA dataset, we conducted a complementary quality assessment using the ROSCOE framework~\cite{golovnevaroscoe}. This evaluation measures step-by-step reasoning coherence across multiple dimensions, including semantic alignment, logical consistency, and contextual grounding. The resulting scores, reported in Table~\ref{tab:roscoe_sa}, further support the reliability and high quality of the generated tables and QA pairs, reinforcing the effectiveness of our data generation pipeline.

\begin{table}[htb]
\centering
\scriptsize
\caption{ROSCOE~\citet{golovnevaroscoe} reasoning metrics averaged over the whole dataset.  The “Direction” column indicates whether higher or lower values correspond to better performance for each metric.}
\begin{tabularx}{0.7\textwidth}{X >{\centering\arraybackslash}p{1.5cm} >{\centering\arraybackslash}p{1.5cm} >{\centering\arraybackslash}p{1.5cm}}
\toprule
\textbf{Metric} & \textbf{Direction} & \textbf{Mean} & \textbf{Std}\\
\midrule
\multicolumn{4}{c}{\textbf{Semantic Adequacy} ($\uparrow$)} \\
\midrule
Faithfulness-Step & $\uparrow$ & 0.99 & 5e-4\\
Informativeness-Step & $\uparrow$ & 0.99 & 5e-3\\
Informativeness-Chain & $\uparrow$ & 0.98 & 1.4e-2\\
Faithfulness-Token & $\uparrow$ & 0.99 & 2e-3 \\
\multicolumn{2}{c}{\textbf{Avg}} & \multicolumn{2}{c}{\textbf{0.99}} \\
\midrule
\multicolumn{4}{c}{\textbf{Redundancy \& Risk} ($\downarrow$)} \\
\midrule
Repetition-Token & $\downarrow$ & 0.06 & 0.12\\
Repetition-Step & $\downarrow$ & 0.06 & 0.12\\
\multicolumn{2}{c}{\textbf{Avg}} & \multicolumn{2}{c}{\textbf{0.06}} \\
\midrule
\multicolumn{4}{c}{\textbf{Logical Inference} ($\uparrow$)} \\
\midrule
Discourse-Representation & $\uparrow$ & 0.68 & 0.41\\
Coherence-Step & $\uparrow$ & 0.7 & 0.40\\
\multicolumn{2}{c}{\textbf{Avg}} & \multicolumn{2}{c}{\textbf{0.69}} \\
\midrule
\multicolumn{4}{c}{\textbf{Fluency \& Perplexity} ($\downarrow$)} \\
\midrule
Perplexity-Step & $\downarrow$ & 0.01 & 0.01\\
Perplexity-Chain & $\downarrow$ & 0.05 & 0.03\\
Perplexity-Step-Max & $\downarrow$ & 8e-3 & 8e-3\\
\multicolumn{2}{c}{\textbf{Avg}} & \multicolumn{2}{c}{\textbf{0.02}} \\
\midrule
\multicolumn{4}{c}{\textbf{Grammaticality} ($\uparrow$)} \\
\midrule
Grammar-Step & $\uparrow$ & 0.96 & 0.05\\
Grammar-Step-Max & $\uparrow$ & 0.9 & 0.13\\
\multicolumn{2}{c}{\textbf{Avg}} & \multicolumn{2}{c}{\textbf{0.93}} \\
\bottomrule
\end{tabularx}
\label{tab:roscoe_sa}
\end{table}

\section{LLM Jury Reliability}
Following recent studies~\citet{hedistill, fu2025refocus, verga2024replacing}, we adopt a high-performing LLM jury combined with a majority-vote strategy to evaluate model predictions. This multi-model jury setup enhances evaluation robustness and mitigates individual model biases. The configuration details for all jury models are provided in Section~\ref{sec:jury-prompt}. As argued in~\citet{verga2024replacing}, aggregating judgments from several strong LLMs yields more consistent and reliable evaluations than relying on a single model. In Section~\ref{sec:llm-vs-human}, we discuss the challenges of LLM-based evaluations and we report a detailed comparison between our LLM jury and human annotators on \texttt{Qwen2.5-VL-7B} predictions across multiple benchmarks. In Section~\ref{sec:llm-juries-agreement}, we provide a detailed analysis of jury-to-jury agreement across both table and QA pair quality assessments, as well as benchmark evaluations. The results reveal strong inter-annotator alignment, validating the consistency and effectiveness of our jury-based evaluation protocol.

\subsection{\label{sec:jury-prompt}LLM Jury Settings}

LLM juries were involved at two key stages of the pipeline: (i) \textbf{quality assurance}, where generated tables and QA pairs were validated before inclusion in the dataset, and (ii) \textbf{evaluation benchmarking}, where model responses were assessed for accuracy and reasoning quality.

The prompt used for quality assurance is shown in Figure~\ref{fig:llm3-prompt}, while the prompt used for evaluation during benchmarking is shown in Figure~\ref{fig:llm4-prompt}. Each jury consisted of multiple high-performing vision-language or reasoning-capable LLMs, and final decisions were made via majority voting.

To ensure consistency and reproducibility across evaluations, all LLM jury calls were executed with a temperature setting between 0.0 and 0.1.

We observed a notable drop in judgment accuracy when LLM juries were instructed to return only a structured JSON verdict without any preceding explanation. In particular, \texttt{Mistral-large} systematically omitted its reasoning whenever the keyword \texttt{JSON} was included in the prompt. This issue was effectively mitigated by explicitly instructing models to provide a rationale prior to their final decision and by avoiding any direct mention of \texttt{JSON} in the prompt, except for \texttt{GPT-4.1} which remained robust under such formatting. Including explicit reasoning significantly improved the reliability and depth of model evaluations, likely by reducing shallow assessments and prompting more thoughtful judgments.

\begin{figure}[htb]
\centering
\begin{minipage}{\textwidth}
\begin{lstlisting}
You are a reasoning question answer expert. You will be given a LaTeX formated table with/without diagram, a list of 3 topics, and a pair of a question and its answer.

Your task is to evaluate the pair of question answer based solely on the data in the LaTeX code and these criteria:

    1) Does the LaTeX code contain a Table (not some charts alone or diagrams alone) ?
    2) Does the table, any optional diagrams, and the rest of the document are on one single topic from the provided list of topics, and internally consistent (be careful to off-topic diagrams)?
    3) Is the question clear and related to the table or the diagram?
    4) Is the answer (including its reasoning) totally valid and does it actually respond to the question?
    5) Is the answer FULLY supported by and ONLY BY the table or diagram data (no extra knowledge)?

If the five criteria are true, mark the pair as correct.
If one of the criteria is not met, mark it as incorrect.

Think step by step and conclude with your decision and the index of the criterium not met (if none, index is 0) as follows:
JSON_mention
{{"decision": [0, index_of_the_criterium_not_met]}} for incorrect or {{"decision": [1, 0]}} for correct.
\end{lstlisting}
\end{minipage}
\caption{LLM prompt used for QA evaluation.}
\label{fig:llm3-prompt}
\end{figure}

\begin{figure}[htb]
\centering
\begin{minipage}{\textwidth}
\begin{lstlisting}
You are an expert evaluator of question-answer pairs. You will be given a question a model's answer and a ground truth answer (reference).
Evaluate the answer based on these criteria:
1) Is the model's answer logically consistent?
2) Does the model's answer convey the same meaning as the ground truth?
If the two criteria are true, mark the pair as correct. If one of the criteria is not met, mark it as incorrect.

Think step by step and conclude with your verdict and the index of the criterium not met (if none, index is 0) as follows:
{{"verdict": [0, index_of_the_criterium_not_met]}} for incorrect or {{"verdict": [1, 0]}} for correct.

Question: {question}
Answer: {answer}
Ground Truth: {ground_truth}
Response:
\end{lstlisting}
\end{minipage}
\caption{LLM prompt used for Benchmark evaluation.}
\label{fig:llm4-prompt}
\end{figure}

\subsection{\label{sec:llm-vs-human}LLM Jury Limitations and Mitigation Strategies}

Our evaluation protocol measures alignment between a model’s prediction and the reference answer, leveraging LLM juries for semantic comparison. This strategy is highly cost-effective, enabling scalable automatic assessments on reasoning tasks. However, it comes with important limitations. A model’s response may rely on external world knowledge rather than explicitly extracting information from the table image. In such cases, LLM juries lack the ability to verify whether the response is grounded in the table itself, as they do not have access to the visual context. This weakness makes the evaluation pipeline vulnerable to false positives, especially when the model outputs a factually correct answer that is not actually derivable from the table content. This could be mitigated by including rendered table images alongside the question and model prediction at evaluation time—giving LLM juries full visual grounding. Unfortunately, While effective, this approach is significantly more costly in terms of API calls and inference latency, making it challenging to scale on large datasets. These observations underscore the need for hybrid evaluation strategies combining automatic LLM-based judgments with human verification, in order to quantify the risk of jury-related errors at scale.

To rigorously assess this risk, we conducted a manual evaluation of \texttt{Qwen2.5-VL-7B-Instruct}'s predictions on both Visual-TableQA and the \textit{VTabFact} split of TableVQA-Bench, one of the benchmarks where the model exhibited degraded performance after fine-tuning. Our manual analysis serves as a ground truth reference to assess the reliability and limitations of LLM-based jury evaluation. Our evaluation considers not only the final answer, but also the validity of the reasoning process: even predictions with the correct answer are marked as incorrect if the chain of thought is flawed.
We used the manual analysis to validate the judgments made by the LLM juries. We identified two types LLM Juries errors:
\begin{itemize}
    \item \textit{False Positive:} The LLM jury mistakenly accepts an incorrect model prediction as correct.
    \item \textit{False Negative:} The LLM jury mistakenly rejects a correct model prediction as incorrect.
\end{itemize}
The errors distribution are detailed in Table~\ref{tab:jury-errors-types}. The results reveal that the LLM jury aligns with human judgment within a reasonable margin of $4.7\%$, even under our strict annotation protocol that penalizes incorrect reasoning regardless of the final answer. A more relaxed evaluation criterion would yield an even smaller discrepancy. These findings reinforce the reliability of our LLM jury setup, demonstrating its effectiveness as a scalable proxy for human evaluation.

\begin{table}[htb]
\centering
\scriptsize
\caption{Misclassifications by LLM juries on Visual-TableQA and \textit{VTabFact}. Percentages are calculated relative to the total number of evaluated examples in each dataset.}
\begin{tabularx}{0.5\textwidth}{X *{2}{>{\centering\arraybackslash}p{2cm}}}
\toprule
\textbf{Error Type} & \textbf{Visual-TableQA} & \textbf{\textit{VtabFact}}\\
\midrule
False Positive & 3.45\% & 1.6\%\\
False Negative & 1.2\% & 0.0\%\\
Total & 4.65\% & 1.6\%\\
\bottomrule
\end{tabularx}
\label{tab:jury-errors-types}
\end{table}

\subsection{\label{sec:llm-juries-agreement}LLM Jury Agreement Analysis}
Figure~\ref{fig:llm-agreement-table} presents a detailed breakdown of agreement levels between individual LLM juries, as well as their alignment with majority-vote annotations. This analysis was conducted across both table and QA pair quality assessments. The analysis reveals a spectrum of consistency across juries, with \texttt{GPT-4.1} emerging as the most reliable, likely due to its robust handling of edge cases. Among all models, proprietary LLMs such as \texttt{Gemini-2.5-pro} and \texttt{GPT-4.1} show the strongest alignment with the majority vote, while \texttt{Deepseek-v3.1} exhibits the weakest agreement. Notably, pairwise jury agreement patterns appear correlated with the models’ reasoning capabilities. Despite variability in alignment strength, all juries demonstrate a meaningful degree of concordance with the majority, underscoring the robustness of our collective evaluation protocol.

Conversely, Figure~\ref{fig:llm-agreement-benchmark} shows consistently strong jury agreement across all models for benchmark evaluations, with no notable divergence between proprietary and open-source LLMs. This can be attributed to the relatively simpler nature of the task (semantic comparison between model predictions and ground truth) compared to the more complex reasoning required for evaluating table-question-answer triples, as analyzed in Figure~\ref{fig:llm-agreement-table}.

\pgfplotsset{compat=1.18}
\usepgfplotslibrary{colorbrewer}

% ---- Table with extra column (majority agreement) ----
\pgfplotstableread{
x y val
1 1 100
2 1 81.9
3 1 79.3
4 1 71.3
5 1 77.7
6 1 87.5
1 2 81.9
2 2 100
3 2 85.8
4 2 70.4
5 2 81.1
6 2 91.7
1 3 79.3
2 3 85.8
3 3 100
4 3 66.1
5 3 83.0
6 3 79.3
1 4 71.3
2 4 70.4
3 4 66.1
4 4 100
5 4 70.8
6 4 74.9
1 5 77.7
2 5 81.1
3 5 83.0
4 5 70.8
5 5 100
6 5 87.8
}\agreesix

\begin{figure}[t]
\centering
\begin{tikzpicture}
\begin{axis}[
  width=\linewidth, height=0.62\linewidth,
  axis on top, enlargelimits=false,
  y dir=reverse, xmin=0.5, xmax=6.5, ymin=0.5, ymax=5.5,
  xtick={1,2,3,4,5,6}, ytick={1,2,3,4,5},
  xticklabels={Gemini-2.5-pro,Gpt-4.1,Mistral-large,Deepseek-v3.1,Deepcogito-v2,Majority},
  yticklabels={Gemini-2.5-pro,Gpt-4.1,Mistral-large,Deepseek-v3.1,Deepcogito-v2},
  tick label style={font=\small},
  colormap/viridis, colorbar,
  point meta min=66, point meta max=100, 
]
  \addplot[
    matrix plot*,
    mesh/cols=6,
    point meta=explicit,
    draw=none,
    nodes near coords={\pgfmathprintnumber[fixed,precision=1]{\pgfplotspointmeta}\%},
    every node near coord/.style={font=\scriptsize, text=white},
  ]
  table[x=x, y=y, meta=val] {\agreesix};
\end{axis}
\end{tikzpicture}
\caption{Pairwise agreement (\%) between LLM juries, plus alignment with majority vote for tables and QAs quality assessment}
\label{fig:llm-agreement-table}
\end{figure}

\pgfplotsset{compat=1.18}
\usepgfplotslibrary{colorbrewer}

% ---- Table with extra column (majority agreement) ----
\pgfplotstableread{
x y val
1 1 100
2 1 94.3
3 1 93.2
4 1 95.9
5 1 93.2
6 1 96.9
1 2 94.3
2 2 100
3 2 93.6
4 2 93.3
5 2 93.4
6 2 96.6
1 3 93.2
2 3 93.6
3 3 100
4 3 93.5
5 3 93.2
6 3 95.8
1 4 95.9
2 4 93.3
3 4 93.5
4 4 100
5 4 93.0
6 4 96.4
1 5 93.2
2 5 93.4
3 5 93.2
4 5 93.0
5 5 100
6 5 95.7
}\agreesix

\begin{figure}[t]
\centering
\begin{tikzpicture}
\begin{axis}[
  width=\linewidth, height=0.62\linewidth,
  axis on top, enlargelimits=false,
  y dir=reverse, xmin=0.5, xmax=6.5, ymin=0.5, ymax=5.5,
  xtick={1,2,3,4,5,6}, ytick={1,2,3,4,5},
  xticklabels={Gemini-2.5-pro,Gpt-4.1,Mistral-large,Deepseek-v3.1,Deepcogito-v2,Majority},
  yticklabels={Gemini-2.5-pro,Gpt-4.1,Mistral-large,Deepseek-v3.1,Deepcogito-v2},
  tick label style={font=\small},
  colormap/viridis, colorbar,
  point meta min=93, point meta max=100, 
]
  \addplot[
    matrix plot*,
    mesh/cols=6,
    point meta=explicit,
    draw=none,
    nodes near coords={\pgfmathprintnumber[fixed,precision=1]{\pgfplotspointmeta}\%},
    every node near coord/.style={font=\scriptsize, text=white},
  ]
  table[x=x, y=y, meta=val] {\agreesix};
\end{axis}
\end{tikzpicture}
\caption{Pairwise agreement (\%) between LLM juries, plus alignment with majority vote for benchmark evaluation}
\label{fig:llm-agreement-benchmark}
\end{figure}

\subsection{Evaluator Consistency Compared to the Literature}

In line with recent studies~\citet{hedistill, fu2025refocus, verga2024replacing, agarwal2025supernova}, we employed a high-performing LLM jury with a majority-vote strategy to evaluate model predictions. The jury consisted of \texttt{Mistral-large}, \texttt{Deepseek-v3.1}, \texttt{Gemini-2.5-pro}, \texttt{GPT-4.1}, and \texttt{Deepcogito-v2}. In Table~\ref{tab:bench-results}, baseline values (in black) are reported from ~\citet{hedistill}, who used \texttt{GPT-4o} as the sole evaluator. While our evaluation pipeline involves a broader and more powerful set of models, making it arguably more reliable and robust ~\citep{verga2024replacing}, we still consider the two evaluation protocols broadly comparable. In fact, due to the stricter majority-vote requirement across diverse models, our approach may even yield more demanding or rigorous evaluations. This comparability also holds for Table~\ref{tab:tableqa_vs_tablevqa}, where baseline performances (in black) are taken from ~\citet{fu2025refocus}, who employed \texttt{GPT-4} for their evaluation. We argue that despite methodological differences, all evaluations are consistent enough to be analyzed jointly for comparative purposes.

\section{\label{sec:hyperparam}Model Finetuning Hyperparameters}

The hyperparameters used for LoRA are reported in Table~\ref{tab:hyperparam}.
For \texttt{Qwen2.5-VL-7B-Instruct}, we employed a two-phase fine-tuning strategy. 
In the first phase (Tier A), we adapted the text-side modules together with the multi-modal projector, while keeping the vision tower frozen. In the second phase (Tier B), we further enabled LoRA on the attention projections of the last four vision blocks, leaving the remaining vision layers untouched.

\begin{table}[htb]
\renewcommand{\arraystretch}{1.3}
\centering
\scriptsize
\caption{Hyperparameters Used for Fine-Tuning with LoRA. More details in our \href{https://github.com/AI-4-Everyone/Visual-TableQA}{GitHub repository}. Abbreviations: lr=learning rate, r= LoRA rank, $\alpha$= LoRA $\alpha$, Targets=targets modules for LoRA.}
\begin{tabularx}{0.65\textwidth}{X *{3}{>{\arraybackslash}p{0.5cm}} >{\arraybackslash}p{2.8cm}}
\toprule
\textbf{Model} & \textbf{lr} & \textbf{r} & \textbf{$\alpha$} & \textbf{Targets}\\
\midrule
LLaVA-Next-Llama3-8B & \multirow{3}{*}{2e-5} & \multirow{2}{*}{16} & \multirow{2}{*}{8} & \multirow{3}{=}{all-linear\\(llm frozen)}\\
MiniCPM-V2.5-Llama3 & & & & \\
InternVL2-8B & & 16$^*$ & 32$^*$ & \\
\multirow{2}{=}{Qwen2.5-VL-7B-Instruct} & 1e-4 & 16 & 8 & Tier A: projector modules$^{**}$\\
& 2e-5 & 8 & 32 & Tier B: last 4 vision blocks’ attention\\
\bottomrule
\multicolumn{5}{p{0.6\textwidth}}{\footnotesize * The InternVL training source code sets the LoRA alpha as twice the LoRA rank, as shown in their official implementation \href{https://github.com/OpenGVLab/InternVL/blob/main/internvl_chat/internvl/train/internvl_chat_finetune.py}{here}. We followed this convention for full reproducibility and assumed that other baselines applied the same rule.}\\

\multicolumn{5}{p{0.6\textwidth}}{\footnotesize ** We left the Vision Tower untouched as it significantly degraded model performance.}
\end{tabularx}
\label{tab:hyperparam}
\end{table}

\section{\label{sec:errors}Error Taxonomy of Model Predictions}

To better understand the failure modes of \texttt{Qwen2.5-VL-7B-Instruct}, we conducted a fine-grained manual analysis of its predictions, both before and after fine-tuning on Visual-TableQA and the \textit{VTabFact} split of TableVQA-Bench. This section is organized as follows:  
\textit{(i)} we define the full set of error types used in our annotation protocol (Section~\ref{sec:error-types});  
%\textit{(ii)} we present a comparative analysis of errors on Visual-TableQA (Section~\ref{sec:visual-errors}); and  
\textit{(ii)} we present a comparative analysis of errors observed on \textit{VTabFact} (Section~\ref{sec:vtabfact-errors}).

\subsection{\label{sec:error-types}Error Categories}
\begin{figure}[htb]
\centering

\begin{subfigure}[t]{0.32\linewidth}
\vspace{0pt}
\centering
\fbox{\includegraphics[width=\linewidth]{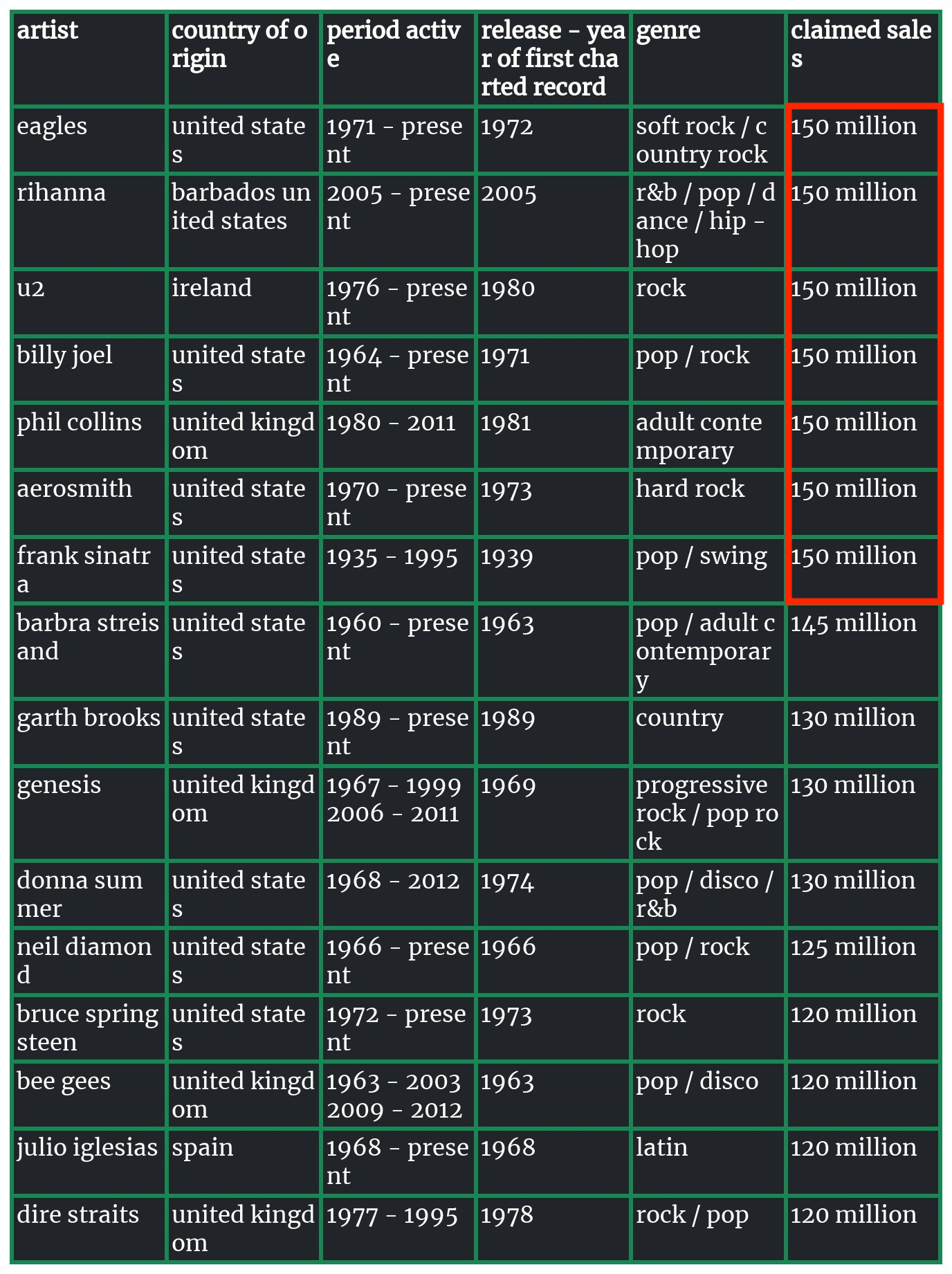}}
\vspace{0.3em}
{\small\textbf{Q:} there be 6 artist with claimed sale of 150 million?}
\begin{tcolorbox}[width=\linewidth, colback=white, colframe=black!30, boxsep=2pt, left=3pt, right=3pt, top=2pt, bottom=2pt]
\scriptsize\textbf{A:} To determine if there are 6 artists with claimed sales of 150 million, let's count the entries in the "claimed sales" column where the value is 150 million:

1. Eagles
[...]
6. Aerosmith

There are indeed 6 artists listed with claimed sales of 150 million.
\end{tcolorbox}
\caption{Partial Data Extraction}
\label{fig:partial}
\end{subfigure}
\hfill%
\begin{subfigure}[t]{0.32\linewidth}
\vspace{0pt}
\centering
\fbox{\includegraphics[width=\linewidth]{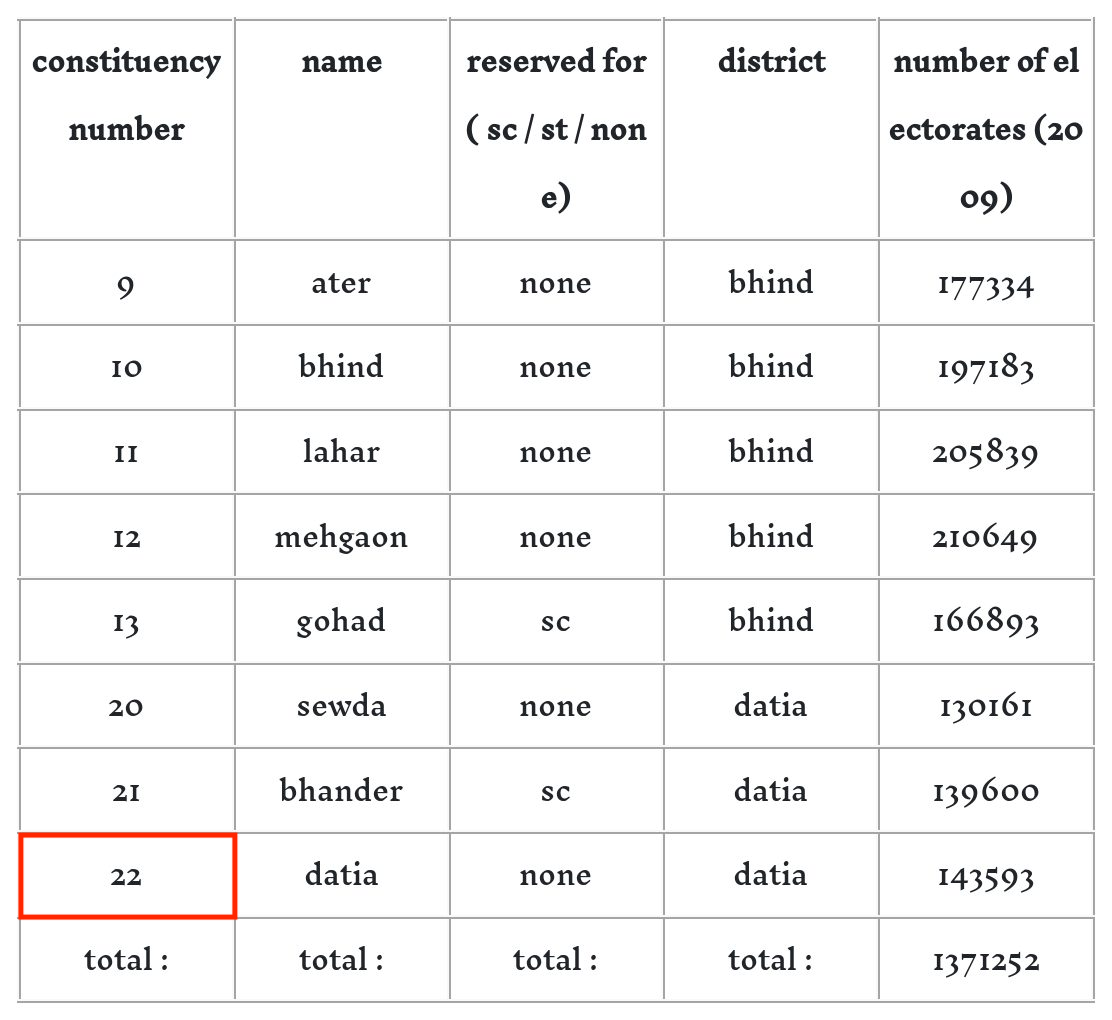}}
\vspace{0.3em}
{\small \textbf{Q:} datia 's constituency number be lower than gohad 's by 10}
\begin{tcolorbox}[width=\linewidth, colback=white, colframe=black!30, boxsep=2pt, left=3pt, right=3pt, top=2pt, bottom=2pt]
\scriptsize\textbf{A:} To determine if Datia's constituency number is lower than Gohad's by 10, we need to compare their respective constituency numbers.

1. Identify the constituency number for Datia: The table shows that Datia's constituency number is 20.
[...]
\end{tcolorbox}
\caption{Hallucination}
\label{fig:hallucination}
\end{subfigure}
\hfill%
\begin{subfigure}[t]{0.32\linewidth}
\vspace{0pt}
\centering
\fbox{\includegraphics[width=\linewidth]{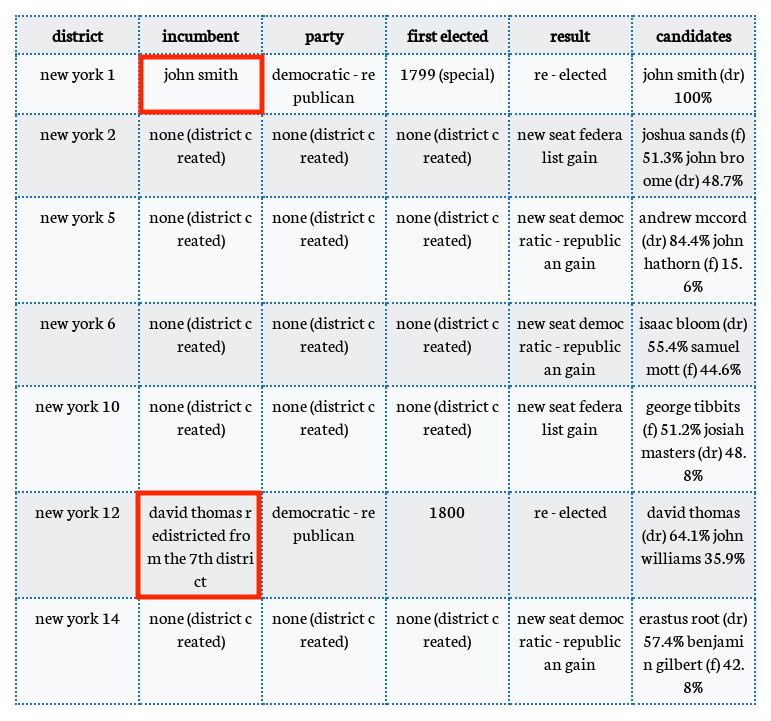}}
\vspace{0.3em}
{\small \textbf{Q:} the majority of the district be new creation and do not have an incumbent}
\begin{tcolorbox}[width=\linewidth, colback=white, colframe=black!30, boxsep=2pt, left=3pt, right=3pt, top=2pt, bottom=2pt]
\scriptsize\textbf{A:} To determine if the majority of districts were newly created and did not have an incumbent, let's analyze the "incumbent" column:

1. **New York 1**: Incumbent: John Smith
[...]
6. **New York 12**: Incumbent: David Thomas
[...]
From this analysis:
- There is only one district (New York 1) that has an incumbent.
[...]
\end{tcolorbox}
\caption{Incoherence}
\label{fig:incoherence}
\end{subfigure}

\vspace{1em}

\begin{subfigure}[t]{0.32\linewidth}
\vspace{0pt}
\centering
\fbox{\includegraphics[width=\linewidth]{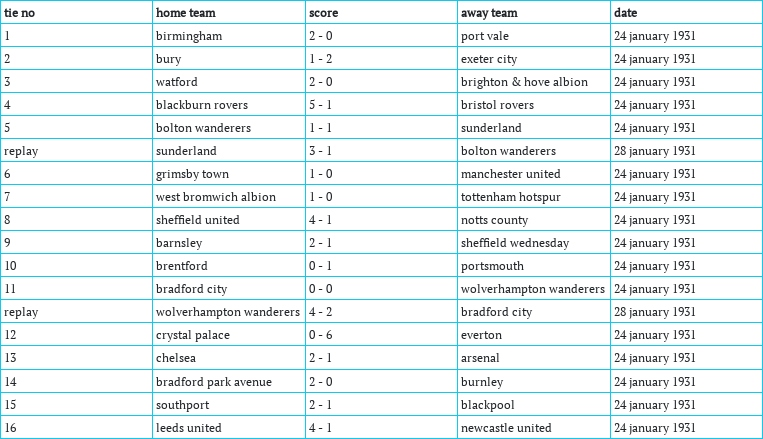}}
\vspace{0.3em}
{\small\textbf{Q:} of the 18 game play , 8 result in 1 or more team be shut out}
\begin{tcolorbox}[width=\linewidth, colback=white, colframe=black!30, boxsep=2pt, left=3pt, right=3pt, top=2pt, bottom=2pt]
\scriptsize\textbf{A:} To determine how many of the 18 games resulted in at least one team being shut out, we need to analyze each game's score:

1. Birmingham vs Port Vale: 2-0 (Birmingham shut out)
2. Bury vs Exeter City: 1-2 (\textcolor{red}{Exeter City shut out})
[...]
\end{tcolorbox}
\caption{Misunderstanding}
\label{fig:misunderstanding}
\end{subfigure}
\hfill%
\begin{subfigure}[t]{0.32\linewidth}
\vspace{0pt}
\centering
\fbox{\includegraphics[width=\linewidth]{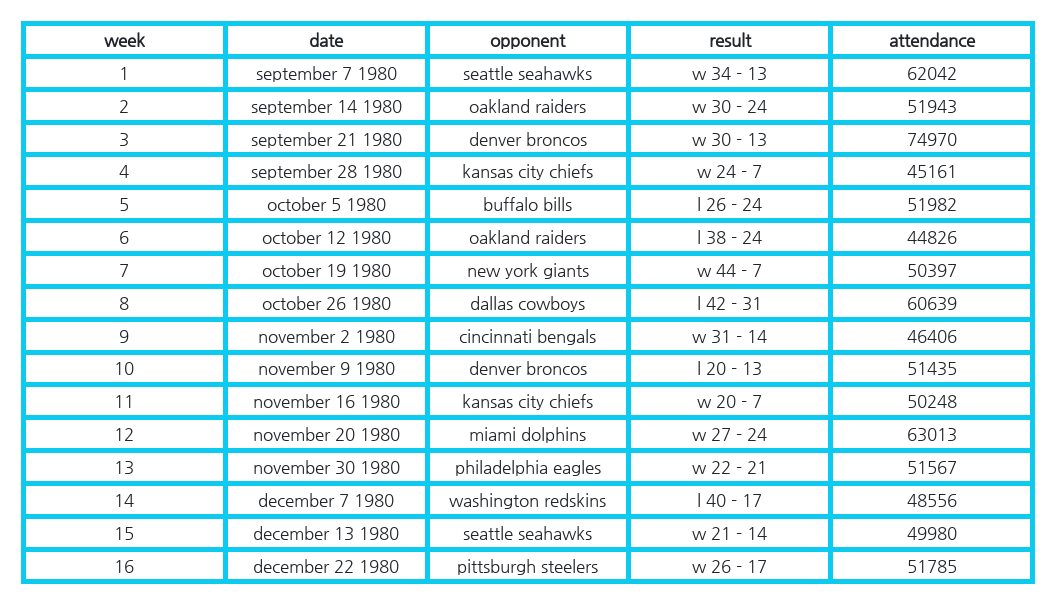}}
\vspace{0.3em}
{\small \textbf{Q:} the smallest crowd the charger play in front of be 44826, a road game against the raider}
\begin{tcolorbox}[width=\linewidth, colback=white, colframe=black!30, boxsep=2pt, left=3pt, right=3pt, top=2pt, bottom=2pt]
\scriptsize\textbf{A:} To determine if the smallest crowd the Chargers played in front of was 44,826 during a road game against the Raiders, let's follow these steps:

1. Identify the games \textcolor{red}{where the Chargers played against the Raiders}:
   [...]
2. Check the attendance for each game:
   [...]
3. Compare the attendance figures: 
[...]
\end{tcolorbox}
\caption{Faulty Methodology}
\label{fig:methodology}
\end{subfigure}
\hfill%
\begin{subfigure}[t]{0.32\linewidth}
\vspace{0pt}
\centering
\fbox{\includegraphics[width=\linewidth]{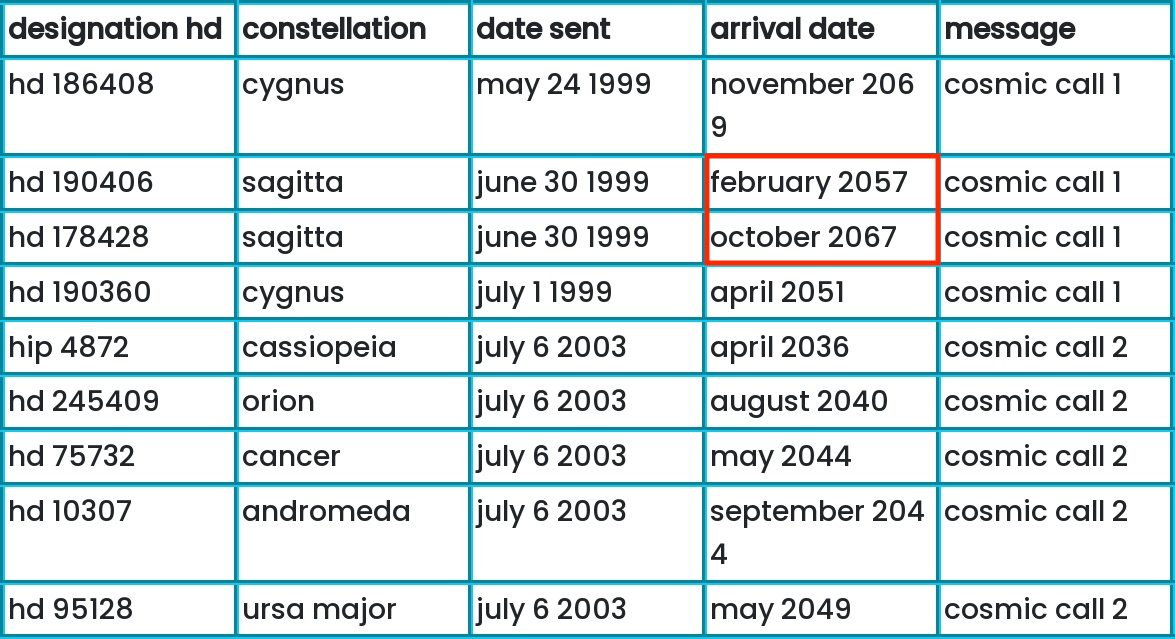}}
\vspace{0.3em}
{\small \textbf{Q:} hd 178428 's arrival date be 10 year later than that of hd 190406}
\begin{tcolorbox}[width=\linewidth, colback=white, colframe=black!30, boxsep=2pt, left=3pt, right=3pt, top=2pt, bottom=2pt]
\scriptsize\textbf{A:} [...]

Since the difference is 10 years and 10 months, it is not exactly 10 years. Therefore, HD 178428's arrival date is not 10 years later than that of HD 190406.

\textbf{Ground Truth:} Yes
\end{tcolorbox}
\caption{Gray Area}
\label{fig:gray}
\end{subfigure}

\caption{Illustration of the Error Categories}
\label{fig:errors-categories}
\end{figure}

We classified the observed errors into eight categories (Figure~\ref{fig:errors-categories}):

\begin{itemize}
    \item \textbf{Partial Data Extraction:} The model overlooks some relevant entries (e.g., stops counting too early, Figure~\ref{fig:partial}).
    \item \textbf{Hallucination:} The model references information not present in the table (Figure~\ref{fig:hallucination}).
    \item \textbf{Incoherence:} The model extracts the correct data but then misinterprets it or contradicts itself later (deductive error, Figure~\ref{fig:incoherence}).
    \item \textbf{Misunderstanding:} The model produces factual statements that do not actually answer the question (Figure~\ref{fig:misunderstanding}).
    \item \textbf{Faulty Methodology/Reasoning:} The reasoning is too shallow, or the model fails to satisfy all the constraints of the query (Figure~\ref{fig:methodology}).
    \item \textbf{Ambiguity (Gray Area):} Both the ground truth and the model’s prediction can be reasonably justified (Figure~\ref{fig:gray}).
    \item \textbf{Dataset Mistake:} The original dataset contains annotation or label errors.
\end{itemize}

%\subsection{\label{sec:visual-errors}Error Analysis on Visual-TableQA}

\subsection{\label{sec:vtabfact-errors}Error Analysis on VTabFact}
\begin{figure}[htb]
    \centering
    % First image
    \begin{subfigure}[t]{0.45\textwidth}
        \centering
        \includegraphics[width=\linewidth]{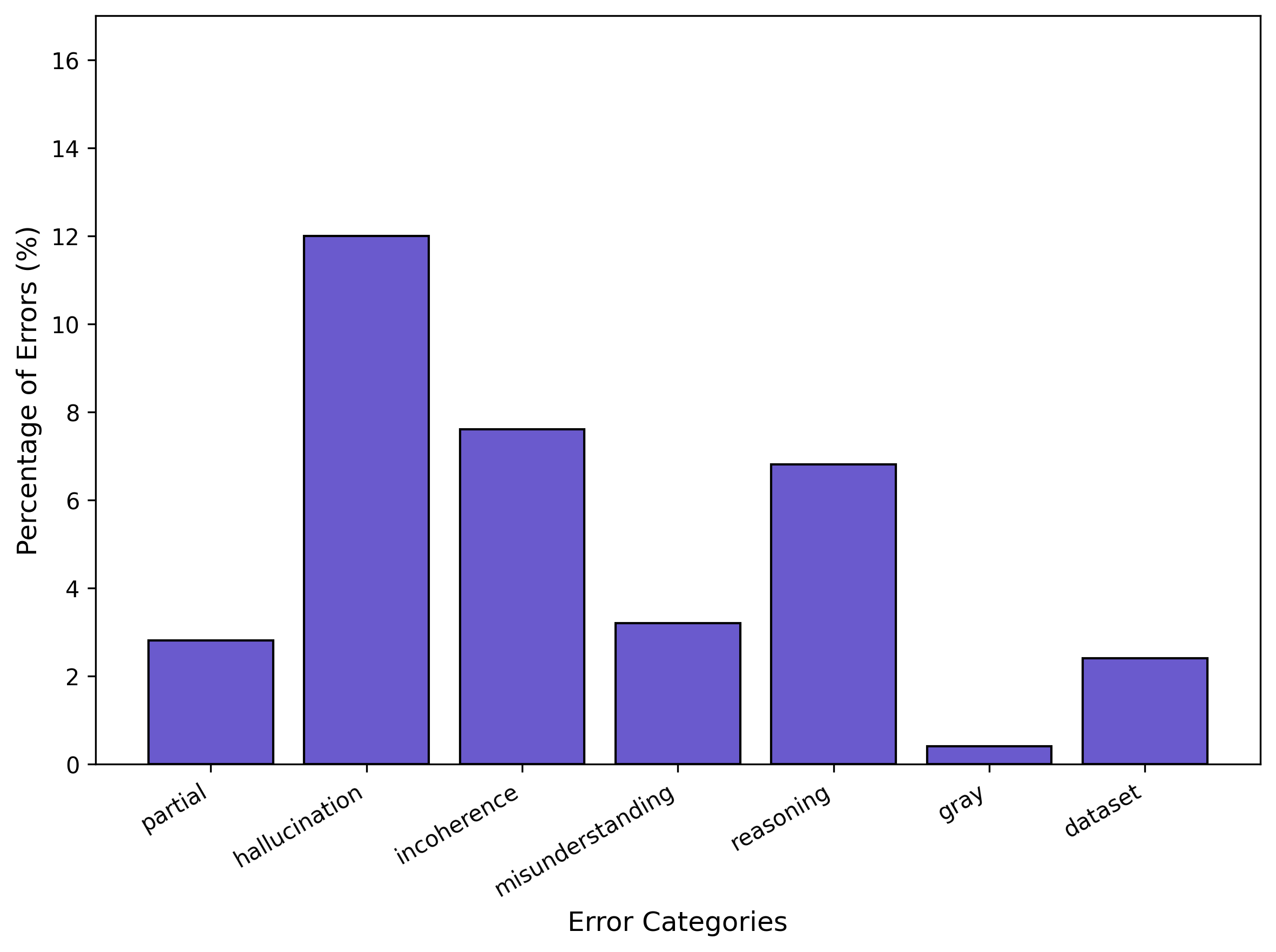}
        \caption{Distribution of Errors for \texttt{Qwen2.5-vl-7b}.}
        \label{fig:vtab-pretrained-errors}
    \end{subfigure}
    \hfill
    % Second image
    \begin{subfigure}[t]{0.45\textwidth}
        \centering
        \includegraphics[width=\linewidth]{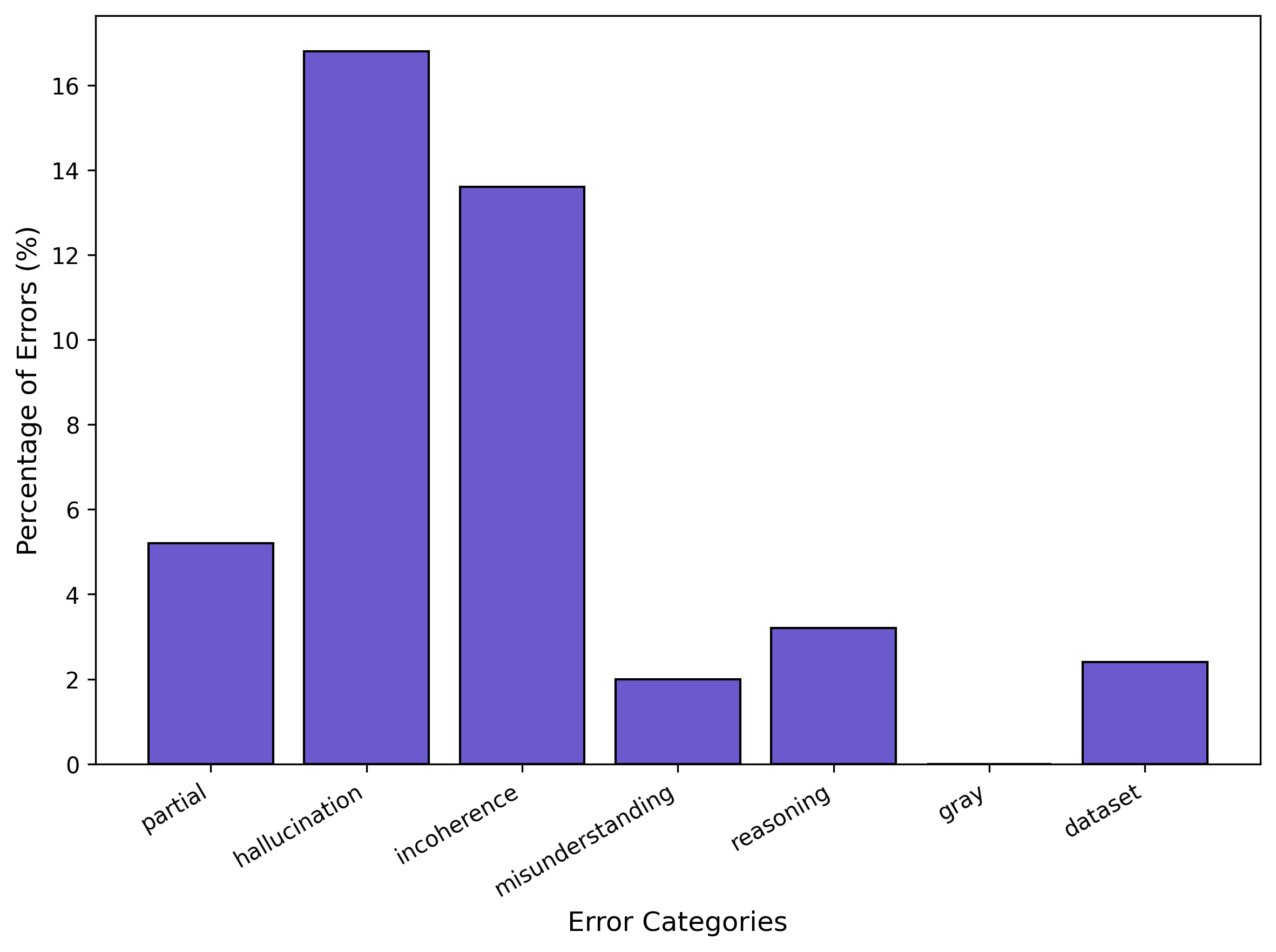}
        \caption{Distribution of Errors for \texttt{Qwen2.5-vl-7b} after Finetuning.}
        \label{fig:vtab-finetuned-errors}
    \end{subfigure}
    % Third image
    \begin{subfigure}[t]{0.45\textwidth}
        \centering
        \includegraphics[width=\linewidth]{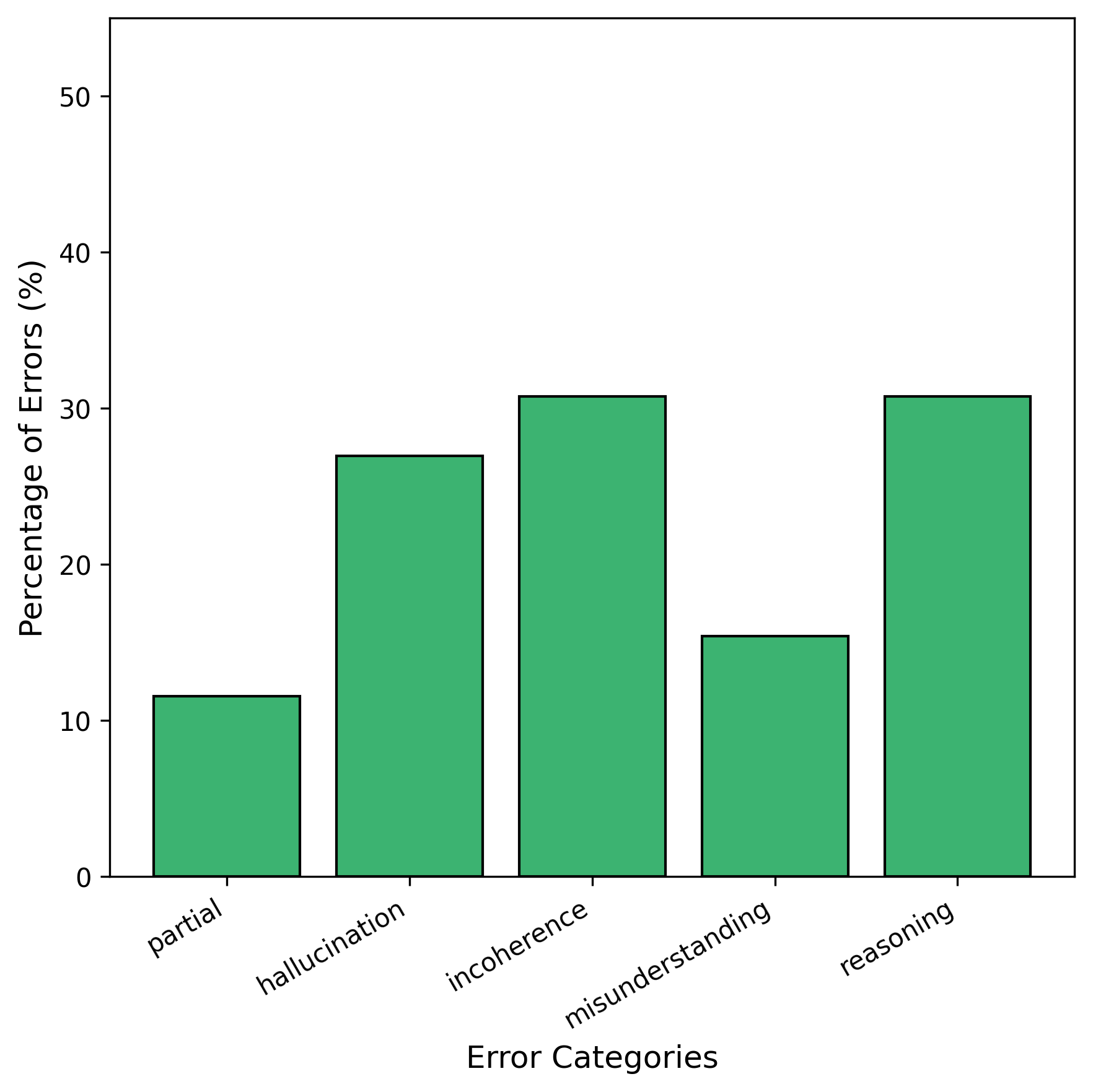}
        \caption{Distribution of Errors corrected by Finetuning.}
        \label{fig:vtab-sft-right-pretrained-wrong}
    \end{subfigure}
    \hfill
    % Second image
    \begin{subfigure}[t]{0.45\textwidth}
        \centering
        \includegraphics[width=\linewidth]{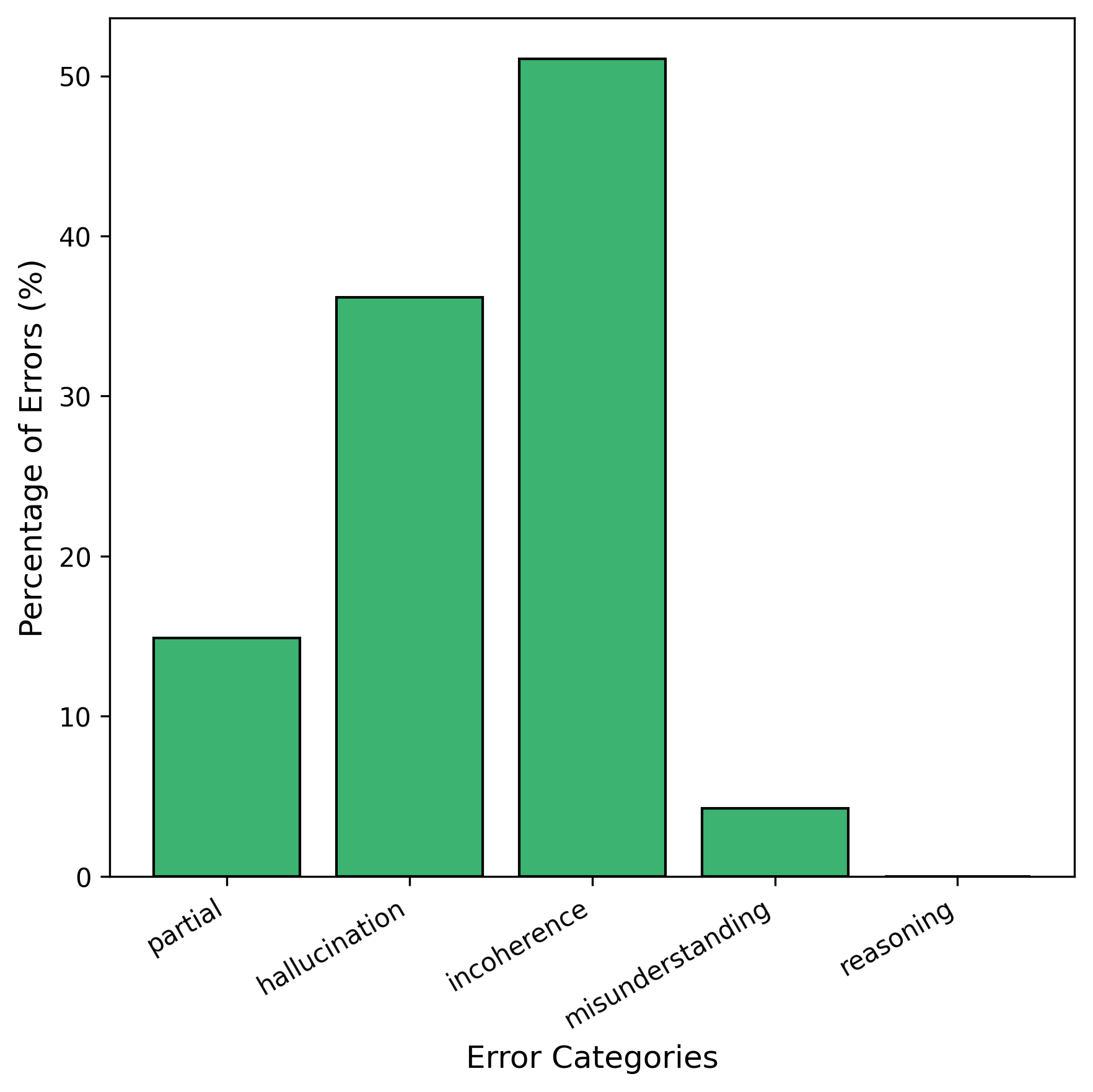}
        \caption{Distribution of Errors introduced after Finetuning.}
        \label{fig:vtab-pretrained-right-sft-wrong}
    \end{subfigure}
    \caption{Comparison of error distributions between the pretrained \texttt{Qwen2.5-VL-7B} model (left) and its finetuned version (right). Percentages in the first row refer to the full \textit{VTabFact} set. Percentages in the second row correspond to: (c) samples where the finetuned model was correct but the pretrained model was not, and (d) samples where the pretrained model was correct but the finetuned model was not.}
    \label{fig:vtab-error-taxonomy}
\end{figure}

The results of our analysis are shown in Figure~\ref{fig:vtab-error-taxonomy}. The side-by-side comparison of Figure~\ref{fig:vtab-pretrained-errors} and Figure~\ref{fig:vtab-finetuned-errors} reveals a significant shift in error distribution after fine-tuning: although the total number of errors increases from 67 to 88 (out of 250 samples), most newly introduced errors concentrate in the \textit{Incoherence} and \textit{Hallucination} categories, while all other error types show a marked decline. In particular, \textit{Faulty Methodology/Reasoning} errors are significantly reduced, indicating that the fine-tuned model exhibits more consistent and structured reasoning patterns. This trend is further supported by Figure~\ref{fig:vtab-sft-right-pretrained-wrong}, which shows the distribution of errors corrected by the fine-tuned model: improvements span across all error categories. In contrast, Figure~\ref{fig:vtab-pretrained-right-sft-wrong} shows that newly introduced errors after fine-tuning are largely concentrated in only a 2 categories. We attribute the sharp rise in \textit{Incoherence} errors to two key factors:

\noindent\textbf{Degraded Arithmetic Capability:} A recurrent issue post-finetuning involves numerical comparison errors or basic math computation errors, leading to faulty deductions despite otherwise correct reasoning steps (see Figure~\ref{}). Such errors were less frequent in the pretrained model.

\noindent\textbf{Answer Template Behavior:} The fine-tuned model tends to state its final answer before providing its chain of thought, often declaring an incorrect answer followed by a valid justification that leads to the correct conclusion (see Figure~\ref{}). Despite the correctness of the rationale, we count such instances as errors. In contrast, the pretrained model typically presents its reasoning first, then concludes—leading to fewer coherence violations.

\section{Visual-TableQA Sample}
Table~\ref{tab:qa-sample} gives some more detailed examples of our dataset samples.

\begin{table}[htb]
\caption{Sample of reasoning-intensive QA pairs. The first row’s question and answer are truncated for readability. These questions address multiple visual aspects and extend beyond simple information extraction to test interpretive reasoning, as illustrated in the second row with a “how” question.}
\centering
\begin{tabularx}{\textwidth}{>{\centering\arraybackslash}p{5cm} @{\hspace{0.75em}}
>{\centering\arraybackslash}p{2.3cm} @{\hspace{0.75em}} 
>{\centering\arraybackslash}p{6.1cm}}
\toprule
\textbf{Table} & \textbf{Question} & \textbf{Answer} \\
\midrule
\end{tabularx}
\scriptsize
\begin{tabularx}{\textwidth}{m{5cm} @{\hspace{0.75em}} m{2.3cm}@{\hspace{0.75em}} m{6.1cm}}
\includegraphics[width=5cm]{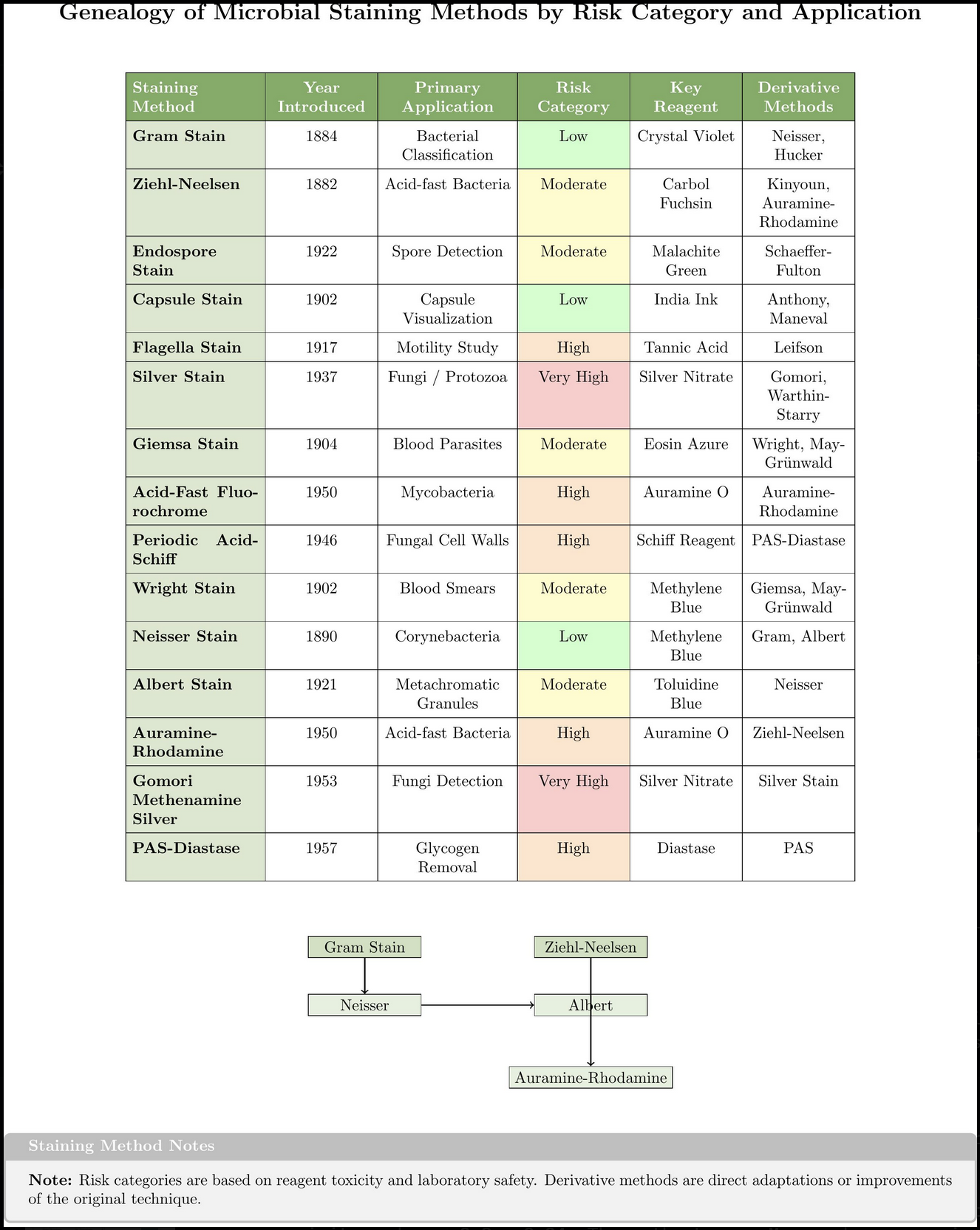} &
Based on the visual patterns and row coloring in the table, what relationship exists between the alternating yellow-highlighted rows and the brewing methods' characteristics [...]?
&
Looking at the table's visual structure, the yellow-highlighted rows (rows 2, 4, 6, and 8) correspond to French Press, Chemex, Espresso, and Cold Brew methods respectively. These methods represent distinct extraction categories: French Press uses immersion, Chemex uses slow filtration, Espresso uses high-pressure extraction, and Cold Brew uses cold water extraction. The alternating color pattern appears to highlight methods that represent fundamental brewing principles or landmark techniques within each difficulty tier.[...]\\
\hline
 \includegraphics[width=5cm]{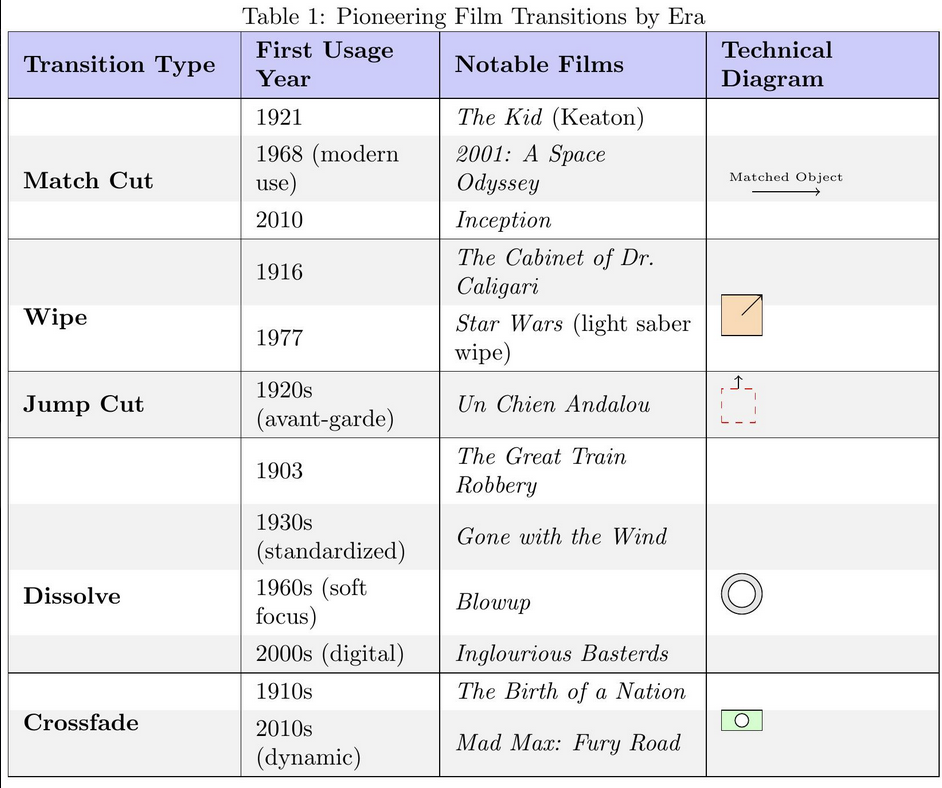}&
 How does the genealogy of origami bases demonstrate a progressive complexity of folding techniques, and what key transformational steps are evident in the progression from a square sheet to more complex bases?
 &
 The genealogy reveals a systematic progression from simple to complex bases, starting with the fundamental Square Sheet. The Preliminary Fold introduces diagonal and perpendicular creases, serving as a critical intermediate step. This base enables more complex transformations like the Kite and Fish Bases, which develop pointed structures. The progression continues to the Bird Base, which combines multiple folding techniques to create more intricate models with distinct appendages. Each base builds upon previous folding techniques, demonstrating an incremental approach to increasing geometric complexity and model potential.\\
 \hline
 \addlinespace[10pt]
 \includegraphics[width=5cm]{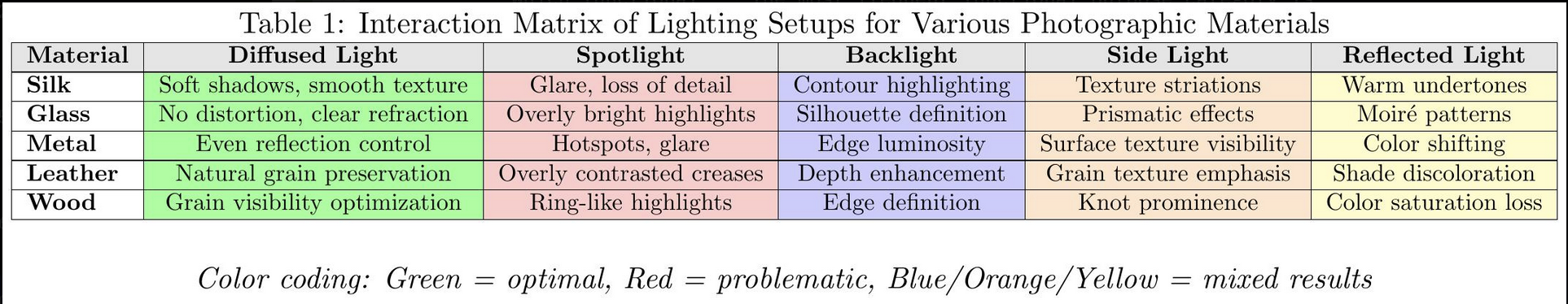}&
 Which UNESCO site has the highest cultural impact indicator, and what symbolic meaning does it represent? How does this relate to its visitor numbers and year of inscription?
 &
 The Great Wall of China has the highest cultural impact indicator with a value of 120. It symbolizes 'Strength and perseverance' (as indicated by the blue color coding and symbolism type 2). Despite having the second-highest visitor numbers (10 million/year), it achieved maximum cultural impact. This suggests that while visitor numbers contribute to cultural impact, other factors like historical significance and symbolic representation play a stronger role. The site was inscribed in 1987, earlier than many others, giving it more time to establish its cultural influence.\\
\bottomrule
\end{tabularx}
\label{tab:qa-sample}
\end{table}

\section{Image-to-LaTeX Dataset}

We have also released an additional dataset, \href{https://huggingface.co/datasets/AI-4-Everyone/Im2TeX}{\textbf{Img2TeX}}, available on Hugging Face. It contains all the table images generated during the construction of \textbf{Visual-TableQA}, along with their corresponding LaTeX source code. This dataset is intended to complement the work of ~\citet{kale2025texpert}, which focuses on evaluating models’ ability to generate LaTeX from textual prompts. In contrast, \textbf{Img2TeX} targets the task of generating LaTeX documents from visual (image-based) inputs. This distinction opens up a new evaluation pathway for vision-language models (VLMs), particularly those aiming to learn structured document generation from visual cues. As such, \textbf{Img2TeX} serves as a valuable benchmark for assessing the visual-to-structured-text generation capabilities of multimodal models.

%%%%%%%%%%%%%%%%%%%%%%%%%%%%%%%%%%%%%%%%%%%%%%%%%%%%%%%%%%%%

\end{document}